\documentclass[11pt]{article}

\usepackage[preprint]{acl}

\usepackage{times}
\usepackage{latexsym}

\usepackage[T1]{fontenc}
\usepackage[utf8]{inputenc}

\usepackage{microtype}
\usepackage{inconsolata}

\usepackage{graphicx}
\usepackage{hyperref}
\usepackage{url}

\usepackage{amsmath,amssymb,amsthm}
\usepackage{amsfonts}
\usepackage{nicefrac}
\usepackage{mathtools}
\usepackage{bbm}

\usepackage{multirow}
\usepackage{makecell}
\usepackage{array}
\usepackage{tabularx}
\usepackage{colortbl}
\usepackage{booktabs}

\usepackage{enumitem}
\usepackage{float}
\usepackage{xspace}

\usepackage{xcolor}
\usepackage{pifont}
\newcommand{\cmark}{\ding{51}}
\newcommand{\xmark}{\ding{55}}

\usepackage{tikz}
\usepackage{pgfplots}
\usepackage{pgfplotstable}
\usepackage{subcaption}
\pgfplotsset{compat=1.18}
\usetikzlibrary{arrows.meta,positioning,shapes.geometric,fit,
                decorations.pathreplacing,calc,backgrounds,matrix}

\usepackage{algorithm,algpseudocode}

\definecolor{clblue}{RGB}{31,119,180}
\definecolor{cloran}{RGB}{214,104,26}
\definecolor{clgrn}{RGB}{44,160,44}
\definecolor{clred}{RGB}{214,39,40}
\definecolor{clpur}{RGB}{148,103,189}
\definecolor{lightgray}{gray}{0.93}
\definecolor{stagecol}{RGB}{219,234,254}
\definecolor{attackcol}{RGB}{254,226,226}
\definecolor{elcol}{RGB}{254,243,199}
\definecolor{eagcol}{RGB}{220,252,231}
\definecolor{mymagenta}{RGB}{205,0,205}

\newcommand{\calG}{\mathcal{G}}

\newcommand{\ELQ}{\textsc{ELQ}\xspace}

\newcommand{\deltaEL}{\Delta_{\mathrm{EL}}}
\newcommand{\deltaSR}{\Delta_{\mathrm{SR}}}

\newcommand{\deltaAG}{\Delta_{\mathrm{AG}}}

\title{Query-Side Attacks on GNN-Based KGQA:\@ Tracing Failures\\
from Entity Linking to Answer Generation}

\author{
  \textbf{Pankaj Kumar}\textsuperscript{1,2}, 
  \textbf{Subhankar Mishra}\textsuperscript{1,2}
\\ 
  \textsuperscript{1}National Institute of Science Education and Research, 
  \textsuperscript{2}Homi Bhabha National Institute \\
  \small{\textbf{Correspondence:} \texttt{\{\{pankaj.kumar,smishra\}@niser.ac.in}}
}

\begin{document}

\maketitle

\begin{abstract}
GNN-based Knowledge Graph Question Answering (KGQA) pipelines process queries through four discrete stages:
entity linking, subgraph retrieval, GNN reasoning, and answer generation.
Standard robustness evaluations conflate stage-level failures into a single
end-to-end metric, obscuring both the source of brittleness and the
appropriate mitigation target. We ask which stage fails, and why, when the pipeline is subjected to adversarial perturbations on the input question.
We introduce a stage-isolation protocol with two answer-preserving
adversarial perturbations verified against the knowledge graph:
Compositional Restructuring (CR) and Relation Synonym Swap (RS) target distinct stages while leaving entity seeds intact. Evaluated across ComplexWebQuestions and WebQSP, the results run counter to prevailing assumptions: the GNN reasoning stage retains near-baseline accuracy when the subgraph is intact, while subgraph construction accounts for over 99\% of the end-to-end collapse under CR, occurring even when the gold answer is present in 74\% of retrieved subgraphs.
This exposes a fundamental distinction between answer presence and answer
reachability that end-to-end metrics cannot detect, and places the
mitigation target firmly at the subgraph construction stage rather than
the reasoning model.
Perturbed datasets and evaluation infrastructure are released at
\url{https://github.com/pankajkumar6002/atkgrag}.
\end{abstract}

\section{Introduction}\label{sec:intro}

Knowledge graph question answering (KGQA) has become a core benchmark for
structured reasoning: given a natural-language question and a large knowledge
graph (KG) such as Freebase, the system must find the answer entity by
traversing multi-hop relational paths~\cite{sun2018open,saxena2020improving,
zhang2022subgraph}. GNN-based pipelines have pushed state-of-the-art accuracy
by coupling a graph neural network (GNN) reasoner with a retrieval-augmented
generation (RAG) framework, achieving strong results on ComplexWebQuestions
(CWQ) and WebQSP~\cite{mavromatis2024gnn}.

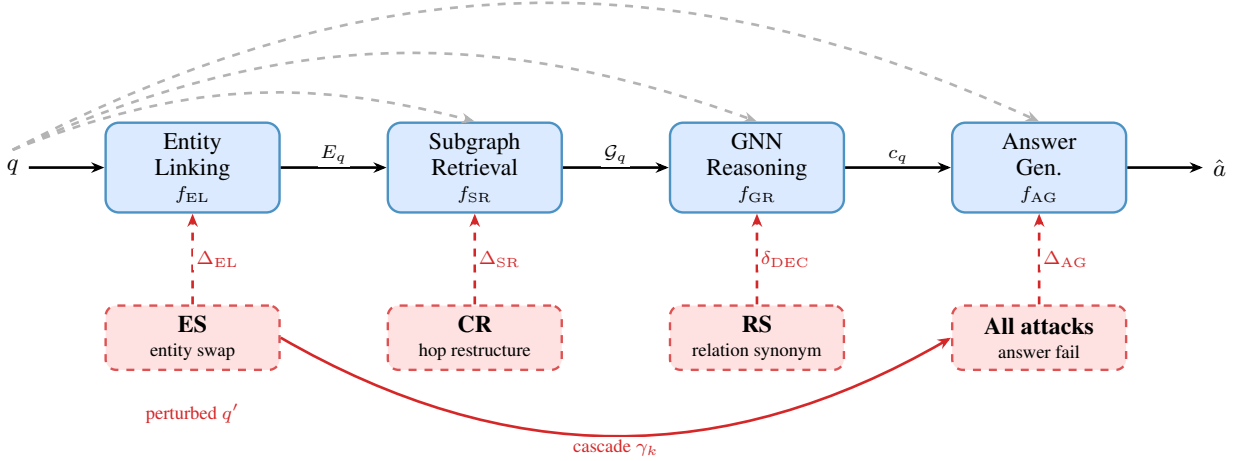
\begin{figure*}[ht]
    \centering
    \begin{tikzpicture}[
        font=\small,
        box/.style={
                draw=clblue!80, fill=stagecol, rounded corners=5pt,
                minimum width=2.3cm, minimum height=1cm,
                align=center,
                line width=0.9pt
            },
        atk/.style={
                draw=clred!80, fill=attackcol, rounded corners=4pt,
                minimum width=2.3cm, minimum height=0.85cm,
                align=center,
                line width=0.9pt, dashed
            },
        arr/.style={-{Stealth[length=5pt,width=4pt]}, line width=1pt},
        darr/.style={-{Stealth[length=5pt,width=4pt]}, line width=1pt,
        clred, dashed},
        lbl/.style={font=\scriptsize, fill=white, inner sep=1pt},
        ]

        \node[box] (EL) {Entity\\Linking\\[-1pt]{\scriptsize $f_{\mathrm{EL}}$}};
        \node[box, right=1.4cm of EL] (SR) {Subgraph\\Retrieval\\[-1pt]{\scriptsize $f_{\mathrm{SR}}$}};
        \node[box, right=1.4cm of SR] (EA) {GNN\\Reasoning\\[-1pt]{\scriptsize $f_{\mathrm{GR}}$}};
        \node[box, right=1.4cm of EA] (AG) {Answer\\Gen.\\[-1pt]{\scriptsize $f_{\mathrm{AG}}$}};

        \draw[arr] (EL.east) -- node[lbl,above] {$E_q$} (SR.west);
        \draw[arr] (SR.east) -- node[lbl,above] {$\mathcal{G}_q$} (EA.west);
        \draw[arr] (EA.east) -- node[lbl,above] {$c_q$} (AG.west);

        \node[left=1cm of EL] (qin) {$q$};
        \node[right=1cm of AG] (aout) {$\hat{a}$};

        \draw[arr] (qin) -- (EL.west);
        \draw[arr] (AG.east) -- (aout);

        \draw[arr, gray!60, dashed, bend left=18] (qin.north) to (SR.north);
        \draw[arr, gray!60, dashed, bend left=22] (qin.north) to (EA.north);
        \draw[arr, gray!60, dashed, bend left=26] (qin.north) to (AG.north);

        \node[atk, below=1.2cm of EL] (a1)
        {\textbf{ES}\\[-1pt]{\scriptsize entity swap}};

        \node[atk, below=1.2cm of SR] (a2)
        {\textbf{CR}\\[-1pt]{\scriptsize hop restructure}};

        \node[atk, below=1.2cm of EA] (a3)
        {\textbf{RS}\\[-1pt]{\scriptsize relation synonym}};

        \node[atk, below=1.2cm of AG] (a4)
        {\textbf{All attacks}\\[-1pt]{\scriptsize answer fail}};

        \draw[darr] (a1.north) -- node[lbl,right] {$\Delta_{\mathrm{EL}}$} (EL.south);
        \draw[darr] (a2.north) -- node[lbl,right] {$\Delta_{\mathrm{SR}}$} (SR.south);
        \draw[darr] (a3.north) -- node[lbl,right] {$\delta_{\mathrm{DEC}}$} (EA.south);
        \draw[darr] (a4.north) -- node[lbl,right] {$\Delta_{\mathrm{AG}}$} (AG.south);

        \draw[arr, clred, bend right=30]
        (a1.east) to node[
            below,
            font=\scriptsize,
            clred,
            fill=white,
            inner sep=1pt
        ] {cascade $\gamma_k$} (a4.west);

        \node[below=0.3cm of a1, font=\scriptsize, clred] {perturbed $q'$};

        \node[
            below right=0.1cm and -0.5cm of AG,
            font=\scriptsize,
            text=gray!70
        ]
        {};

    \end{tikzpicture}
\caption{\textbf{GNN-RAG pipeline and perturbation entry points.}
  The pipeline ($f_{\mathrm{EL}} \to f_{\mathrm{SR}} \to f_{\mathrm{GR}}
  \to f_{\mathrm{AG}}$) maps question $q$ to answer $\hat{a}$.
  \textbf{ES} targets $f_{\mathrm{EL}}$ (entity swap $\Rightarrow$ ELQ seed failure).
  \textbf{CR} targets $f_{\mathrm{SR}}$ (hop-order reversal $\Rightarrow$ topology mismatch).
  \textbf{RS} targets $f_{\mathrm{GR}}$ (predicate synonym $\Rightarrow$ instruction drift).
  Dashed grey arrows: $q$ feeds all stages directly.}\label{fig:pipeline_overview}
\end{figure*}

The path from question to answer runs through four discrete stages: Entity
Linking, Subgraph Retrieval, GNN reasoning, and Answer generation.  Each stage
can fail independently and pass corrupted outputs to the next.  An adversary
who reformulates the query without touching the KG or model can trigger a
cascade of failures through this pipeline. Figure~\ref{fig:pipeline_overview}
shows the four stages and their distinct attack surfaces.

Most robustness work on RAG focuses on \emph{corpus-side} attacks: poisoning
the retrieval index~\cite{zhong2023poisoning,zou2024poisonedrag,chaudhari2024phantom},
injecting adversarial passages~\cite{perez2022ignore,shi2023large}, or
manipulating retrieved context~\cite{xue2024badrag}.  These require the
adversary to write to the KG or corpus, a strong and often unrealistic
capability. Existing KGQA robustness studies~\cite{percin-etal-2025-investigating} have explored entity surface-form
noise and relation paraphrase perturbations but have not measured \emph{which
pipeline stage fails first}, nor how that failure propagates downstream.

We conduct the first systematic, stage-resolved robustness study of a
GNN-based KGQA pipeline. The adversary needs only the ability to submit a
question, with no KG access or model internals. To attribute failures to the
correct stage, we introduce a \emph{stage-isolation protocol}: fix upstream
outputs, perturb only the query, and measure per-stage degradation separately.
Two attacks probe distinct stages: \textbf{Compositional Restructuring (CR)}
rewrites the question via hop-order reversal or constraint injection, leaving
entity mentions intact and preserving the SPARQL answer; \textbf{Relation
Synonym Swap (RS)} replaces predicate surface forms with synonyms while keeping
entity mentions unchanged.  Entity Swap (ES) is a diagnostic perturbation
(Appendix~\ref{app:es_diagnostic}) that changes the gold answer by design and
characterises entity-linker brittleness.

{Under CR, subgraphs built from the ELQ entity linker~\cite{li2020efficient}} contain the gold answer in 74\% of cases yet GNN-RAG achieves only 0.68\% EM {(Exact Match)} on CWQ, the answer is present but {Personalized-PageRank (PPR) retrieval topology} is anchored to the original reasoning chain.  GraftNet's question-embedding-weighted PPR achieves 29.82\% EM at 63.3\% answer coverage, higher EM than ELQ's 0.68\% despite \emph{lower} coverage (63.3\% vs.\ 74\%), because it reorients PPR mass
toward the restructured reasoning chain rather than the original one.
Under RS the ELQ subgraph is structurally unchanged, so RS ELQ EM
(20.3\% CWQ, 50.9\% WebQSP) measures GNN instruction decoder sensitivity in
near-isolation.  Path injection at inference recovers 51.4\% CWQ EM under CR
without fine-tuning, matching or exceeding the relation-path-augmented model.
EPR-KGQA (single-shot pattern-based retrieval) retains near-baseline accuracy
under both attacks; ExplaiGNN (iterative pattern-matching, Wikidata) collapses
like GNN-RAG, showing robustness requires single-shot rather than PPR-free retrieval~\cite{EPR-KGQA,explaignn2022}.

\textbf{Contributions.}
\begin{enumerate}[leftmargin=1.5em,itemsep=2pt]
  \item \textbf{Stage-isolation framework.}
        A protocol and two adversarial perturbation types (CR, RS) that probe
        distinct pipeline stages, enabling per-stage degradation measurement
        ($\deltaEL$, $\deltaSR$, $\deltaAG$ {defined in \S\ref{sec:metrics}}) on standard KGQA benchmarks.

  \item \textbf{Answer presence $\neq$ answer reachability.}
        Under CR, 74\% subgraph answer presence yields only 0.68\% EM (CWQ);
        GraftNet ({GEM: Golden Entity Map} oracle seeds) at 63.3\% coverage reaches 29.82\% EM because
        its PPR topology aligns with the restructured reasoning chain.

  \item \textbf{Subgraph construction is the performance ceiling.}
        {Stage-isolation attributes 52.08 percentage points (pp) of} the 52.22\,pp total CR CWQ drop
        to subgraph topology failure; the GNN instruction decoder is robust when
        the subgraph is intact (52.76\% CWQ EM on clean subgraph + CR question).

  \item \textbf{Inference-time path injection recovers most of the gain.}
        Injecting predicted relation paths at inference (no fine-tuning) recovers
        51.4\% CWQ EM under CR, matching or exceeding relation-path-augmented fine-tuning.
\end{enumerate}

\section{Related Work}\label{sec:rw}

\paragraph{GNN-based KGQA and subgraph retrieval.}
Semantic parsing methods~\cite{berant2013semantic,yih2015semantic,lan2021complex}
translate questions into SPARQL for direct KG execution.
Embedding-based methods~\cite{saxena2020improving,zhang2022subgraph,he2021improving}
retrieve answers via dense similarity.
GNN-RAG~\cite{mavromatis2024gnn} combines both: a three-layer GAT (ReaRev)~\cite{mavromatis2022rearev}
over a PPR-retrieved subgraph, with ELQ as entity linker, creating a cascading
failure structure that we characterise under query perturbation.
Subgraph retrieval variants include GraftNet-Orig~\cite{sun2018open} (PPR from
seed entities), NSM~\cite{he2021improving} (relation-weighted BFS pruning), and
EPR-KGQA~\cite{EPR-KGQA} (atomic adjacency patterns, state-of-the-art on CWQ as
of 2024 and near-baseline-retaining under our perturbations,
Section~\ref{sec:exp}).
{Recent learned retrievers address the same subgraph bottleneck through
differentiable or LM-based selection, faithful reasoning over retrieved
evidence, or PPR combined with synonym
links~\cite{huang-etal-2024-less-GSR,gao-etal-2025-d-rag,sui-etal-2025-fidelis,Hippo-RAG}.}

\paragraph{Entity linking and query-side robustness.}
ELQ~\cite{li2020efficient} (BERT-Large bi-encoder) is GNN-RAG's production
linker; BLINK~\cite{wu2020scalable} and ReFinED~\cite{ayoola2022refined}
offer stronger alternatives, and recent evaluations~\cite{li-etal-2025-leveraging-power-Entity-Linking,hou2025harnessingdeepllmparticipation-entity-link,Entity-Linking-Eval}
show LLM-based linkers substantially outperform bi-encoder approaches.
For our two primary attacks (CR and RS), entity linking is intact
(${\leq}0.3$\,pp SeedHit drop), so the subgraph topology is the bottleneck,
not the linker.
BYOKG-RAG~\cite{mavromatis2025byokgragmultistrategygraphretrieval} proposes
iterative LLM-based EL; its improvements are complementary to our structural
findings.
Query-level attacks on text classifiers (TextFooler~\cite{jin2020bert},
BERT-Attack~\cite{li2020bert}) and KGQA
perturbations~\cite{percin-etal-2025-investigating} have studied surface-form
noise, but none measure \emph{which pipeline stage fails first} or how failure
propagates downstream, which is the gap our stage-isolation protocol fills.
Corpus-side attacks (PoisonedRAG~\cite{zou2024poisonedrag},
\citealt{zhong2023poisoning,chaudhari2024phantom}) require KG write access;
our threat model requires only the ability to submit a question.
Concurrently, \citet{zhou2026breaksknowledgegraphbased} find missing
intermediate-hop triples as the dominant failure, validating our subgraph
bottleneck from the corpus side; \citet{ma2026llmgnn} study the inverted
setting where LLM node features absorb structural perturbations, unlike our
pipeline where entity-set errors propagate through message-passing.
\citet{kandpal2023large} show LLMs fail on rare entities, consistent with our
ES diagnostic; \citet{zhao2025ragsafety} provide stage-wise KG-poisoning
analysis complementary to our query-side isolation.

\section{Methodology}\label{sec:method}

\subsection{Pipeline Formalisation}

We model GNN-RAG as a four-stage function composition:
\begin{equation}
  q \;\xrightarrow{f_{\mathrm{EL}}}\; E_q
    \;\xrightarrow{f_{\mathrm{SR}}}\; \calG_q
    \;\xrightarrow{f_{\mathrm{GR}}}\; c_q
    \;\xrightarrow{f_{\mathrm{AG}}}\; \hat{a}
  \label{eq:pipeline}
\end{equation}
where $E_q$ is the linked entity set of Freebase {\emph{machine identifiers} (MIDs; e.g., \texttt{m.02mjmr})}, $\calG_q$ is the retrieved
answer subgraph, $c_q$ is the verbalised evidence context, and $\hat{a}$ is the
generated answer.  In GNN-RAG: $f_{\mathrm{EL}}$ is
\ELQ{}~\cite{li2020efficient}; $f_{\mathrm{SR}}$ is personalised PageRank (PPR)
over the multi-hop neighbourhood seeded by $E_q$~\cite{he2021improving};
$f_{\mathrm{GR}}$ is a three-layer GAT (ReaRev backbone)~\cite{mavromatis2024gnn};
and $f_{\mathrm{AG}}$ is fine-tuned Llama-2-7B.

\subsection{Threat Model}\label{sec:threat}

\textbf{Attacker capability.}
The adversary (i)~reads $q$ before submission; (ii)~submits $q'$ and observes
$\hat{a}$; (iii)~runs local copies of open-source components (perturbation
generator, Freebase SPARQL endpoint) to verify SPARQL denotation preservation.
No access to model parameters or intermediate states is required.  This is a
\emph{query-only black-box} threat model.

\textbf{Semantic validity budget.}
{Every CR/RS perturbation must satisfy: (i)~perplexity $\mathrm{PPL}(q') < 50$}
(GPT-2-large); (ii)~$\mathrm{BERTScore}(q, q') > 0.85$ (DeBERTa-xlarge-mnli);
(iii)~$\mathrm{den}(q', \mathrm{Freebase}) = \mathrm{den}(q, \mathrm{Freebase})$
verified via local Virtuoso.  Formal perturbation rules are in
Appendix~\ref{app:perturbation_rules}.
{The perturbation generator and the SPARQL endpoint constitute offline
evaluation infrastructure rather than attacker capability.  They certify that
a rewritten question preserves the gold answer, and this certification is what
allows a measured accuracy drop to be attributed to the system rather than to
altered question semantics.  The adversary does not require them at attack
time.  The deployment-time attack consists of submitting a single reworded
question.}

\subsection{Perturbation Taxonomy}\label{sec:taxonomy}

We define two primary adversarial attacks plus a diagnostic perturbation
(ES, Appendix~\ref{app:es_diagnostic}).
Four secondary types (S1--S4) are in Appendix~\ref{app:ablation_full}.

\begin{description}[leftmargin=1.5em,itemsep=4pt]

  \item[Entity Swap (ES): diagnostic only.] Substitutes the topic entity,
    changing both the Freebase MID {(Machine-ID)} and the correct answer.  Not an adversarial
    attack; measures entity-linker brittleness.  Results in
    Appendix~\ref{app:es_diagnostic}.

  \item[Compositional Restructuring (CR).] Apply exactly one KG-safe structural
    operation: hop-order reversal, distractor constraint injection, or
    intermediate-entity alias substitution.  Entity mentions are unchanged;
    SPARQL denotation is preserved.  \emph{Pipeline target}: $f_{\mathrm{SR}}$
    and GNN multi-hop traversal.

  \item[Relation Synonym Swap (RS).] Replace the predicate phrase with a
    synonym preserving the underlying Freebase relation.  All entity mentions
    remain exactly unchanged.  \emph{Pipeline target}: GNN instruction decoder.
    Because entity seeds and PPR walks are unchanged, RS ELQ EM numbers
    (20.3\% CWQ, 50.9\% WebQSP) measure GNN instruction decoder sensitivity
    in near-isolation.

\end{description}

All perturbation types are generated by Llama-3.3-70B-Instruct
(prompts in Appendix~\ref{app:prompts}).
Pass rates: 92.4\% on filters~(i)--(ii) and 94.1\% on the SPARQL
denotation check. Perturbation quality was further validated by manual inspection of 500 random samples per attack type (excluding ES), confirming semantic naturalness and answer preservation.

\subsection{Subgraph Retrieval Variants}\label{sec:sr_variants}

All variants use the same GNN-RAG ReaRev checkpoint; only $\calG_{q'}$ changes,
isolating subgraph construction quality from GNN model quality.
Full descriptions of ELQ, GraftNet-Orig, NSM, GraftNet (ours), and EPR-KGQA
are in Appendix~\ref{app:sr_variants}.

\section{Stage-Wise Evaluation Framework}\label{sec:eval}
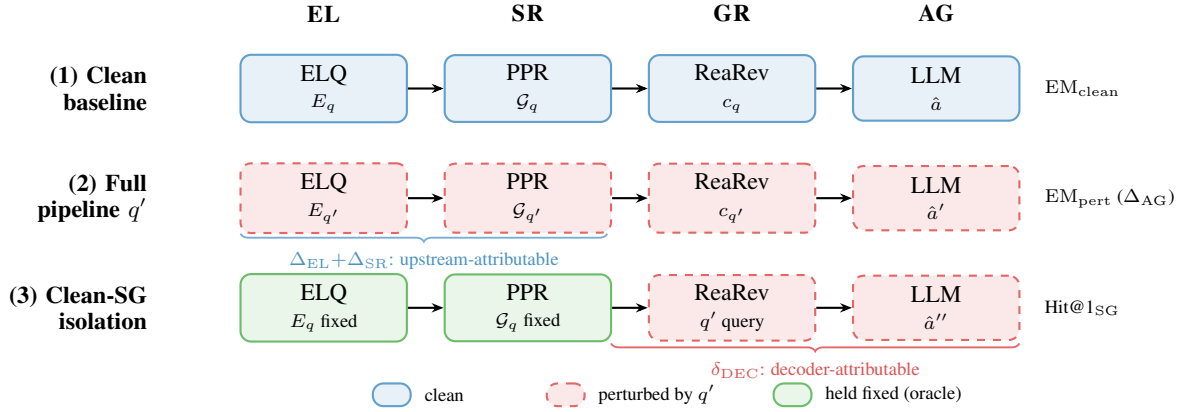
\begin{figure*}[t]
\centering
\begin{tikzpicture}[
    font=\small,
    clean/.style={
        draw=clblue!70, fill=clblue!10, rounded corners=4pt,
        minimum width=2.2cm, minimum height=0.85cm,
        align=center, line width=0.8pt
    },
    pert/.style={
        draw=clred!70, fill=clred!10, rounded corners=4pt,
        minimum width=2.2cm, minimum height=0.85cm,
        align=center, line width=0.8pt, dashed
    },
    fix/.style={
        draw=clgrn!70, fill=clgrn!10, rounded corners=4pt,
        minimum width=2.2cm, minimum height=0.85cm,
        align=center, line width=0.8pt
    },
    arr/.style={-{Stealth[length=4.5pt,width=3.5pt]}, line width=0.8pt},
    rowlbl/.style={font=\footnotesize\bfseries, anchor=east, text width=2.2cm, align=right},
    collbl/.style={font=\footnotesize\bfseries, anchor=south},
]

\def\xEL{2.3}
\def\xSR{5.0}
\def\xGR{7.7}
\def\xAG{10.4}
\def\xOUT{11.7}

\node[collbl] at (\xEL, 5.0) {\textsc{EL}};
\node[collbl] at (\xSR, 5.0) {\textsc{SR}};
\node[collbl] at (\xGR, 5.0) {\textsc{GR}};
\node[collbl] at (\xAG, 5.0) {\textsc{AG}};

\def\yA{4.25}
\node[rowlbl] at (0.1, \yA) {(1) Clean\\baseline};
\node[clean] (r1el) at (\xEL, \yA) {ELQ\\{\scriptsize $E_q$}};
\node[clean] (r1sr) at (\xSR, \yA) {PPR\\{\scriptsize $\mathcal{G}_q$}};
\node[clean] (r1gr) at (\xGR, \yA) {ReaRev\\{\scriptsize $c_q$}};
\node[clean] (r1ag) at (\xAG, \yA) {LLM\\{\scriptsize $\hat{a}$}};
\draw[arr] (r1el.east) -- (r1sr.west);
\draw[arr] (r1sr.east) -- (r1gr.west);
\draw[arr] (r1gr.east) -- (r1ag.west);
\node[font=\scriptsize, anchor=west] at (\xOUT, \yA) {$\mathrm{EM}_{\mathrm{clean}}$};

\def\yB{2.80}
\node[rowlbl] at (0.1, \yB) {(2) Full\\pipeline~$q'$};
\node[pert] (r2el) at (\xEL, \yB) {ELQ\\{\scriptsize $E_{q'}$}};
\node[pert] (r2sr) at (\xSR, \yB) {PPR\\{\scriptsize $\mathcal{G}_{q'}$}};
\node[pert] (r2gr) at (\xGR, \yB) {ReaRev\\{\scriptsize $c_{q'}$}};
\node[pert] (r2ag) at (\xAG, \yB) {LLM\\{\scriptsize $\hat{a}'$}};
\draw[arr] (r2el.east) -- (r2sr.west);
\draw[arr] (r2sr.east) -- (r2gr.west);
\draw[arr] (r2gr.east) -- (r2ag.west);
\node[font=\scriptsize, anchor=west] at (\xOUT, \yB) {$\mathrm{EM}_{\mathrm{pert}}$~($\deltaAG$)};

\draw[decorate, decoration={brace, amplitude=3pt, mirror},
      clblue!60, line width=0.7pt]
  (1.2, 2.33) -- (6.05, 2.33)
  node[midway, below=3pt, font=\scriptsize, clblue!80]
  {$\deltaEL{+}\deltaSR$: upstream-attributable};

\def\yC{1.35}
\node[rowlbl] at (0.1, \yC) {(3) Clean-SG\\isolation};
\node[fix]  (r3el) at (\xEL, \yC) {ELQ\\{\scriptsize $E_q$ fixed}};
\node[fix]  (r3sr) at (\xSR, \yC) {PPR\\{\scriptsize $\mathcal{G}_q$ fixed}};
\node[pert] (r3gr) at (\xGR, \yC) {ReaRev\\{\scriptsize $q'$ query}};
\node[pert] (r3ag) at (\xAG, \yC) {LLM\\{\scriptsize $\hat{a}''$}};
\draw[arr] (r3el.east) -- (r3sr.west);
\draw[arr] (r3sr.east) -- (r3gr.west);
\draw[arr] (r3gr.east) -- (r3ag.west);
\node[font=\scriptsize, anchor=west] at (\xOUT, \yC) {Hit@1$_{\mathrm{SG}}$};

\draw[decorate, decoration={brace, amplitude=3pt, mirror},
      clred!70, line width=0.7pt]
  (6.1, 0.88) -- (11.5, 0.88)
  node[midway, below=3pt, font=\scriptsize, clred!80]
  {$\delta_{\mathrm{DEC}}$: decoder-attributable};

\def\yleg{0.20}
\node[clean, minimum width=0.5cm, minimum height=0.3cm] at (3.2, \yleg) {};
\node[font=\scriptsize, anchor=west] at (3.5, \yleg) {clean};
\node[pert,  minimum width=0.5cm, minimum height=0.3cm] at (5.5, \yleg) {};
\node[font=\scriptsize, anchor=west] at (5.8, \yleg) {perturbed by $q'$};
\node[fix,   minimum width=0.5cm, minimum height=0.3cm] at (8.5, \yleg) {};
\node[font=\scriptsize, anchor=west] at (8.8, \yleg) {held fixed (oracle)};

\end{tikzpicture}
\caption{\textbf{Stage-isolation protocol (three configurations).}
  \textbf{(1) Clean baseline}: question $q$ go through all stages and reference $\mathrm{EM}_{\mathrm{clean}}$.
  \textbf{(2) Full pipeline}: perturbed $q'$ through all stages, which measures total drop $\deltaAG$.
  \textbf{(3) Clean-SG isolation}: EL and SR frozen at clean outputs; only the question text fed to GR is replaced with $q'$.
  Gap $(2){-}(3)$ = subgraph-attributable failure $\delta_{\mathrm{SG}}$;
  gap $(1){-}(3)$ = decoder-attributable failure $\delta_{\mathrm{DEC}}$.}\label{fig:stage_isolation}
\end{figure*}

\subsection{Per-Stage Degradation Metrics}\label{sec:metrics}

We track failure propagation through four metrics ($\deltaEL$, $\deltaSR$,
$\alpha_{\mathrm{pert}}$, $\deltaAG$), one per pipeline stage: 

\textbf{Entity linking degradation $(\Delta_{EL})$}
\begin{equation}
  \deltaEL = \mathrm{SeedHit}(E_q) - \mathrm{SeedHit}(E_{q'})
\end{equation}
where
\begin{equation}
  \mathrm{SeedHit}(E) = \frac{1}{N}\sum_{i=1}^{N}
    \mathbf{1}\!\left[\hat{E}_i \cap E_i^{*} \neq \emptyset\right]
\end{equation}
$\hat{E}_i$ is the set of MIDs linked by ELQ for question $i$;
$E_i^{*}$ is the set of gold seed MIDs.  We use any-match because GNN-RAG
requires only one correct seed entity to initiate PPR.

\textbf{Subgraph retrieval drift $(\deltaSR)$}
\begin{equation}
  \deltaSR = 1 - J(\calG_q,\, \calG_{q'})
\end{equation}
where $J$ is the triple-level Jaccard on Freebase string triples $(s,p,o)$.
High $\deltaSR$ means the subgraph has changed substantially;
$\alpha_{\mathrm{pert}}$ (below) measures quality directly.

{We use a triple-level Jaccard for three reasons: (i) it measures the
\emph{structural} drift of the retrieved evidence, complementing
$\alpha_{\mathrm{pert}}$, which measures answer \emph{presence}; (ii) matching over
string triples is exact, so no embedding model or similarity threshold has to
be tuned; and (iii) it is the natural set-overlap measure over the subgraph object
that the pipeline actually passes downstream.  No character $n$-gram
similarity is involved.}

\textbf{$\alpha_{\mathrm{pert}}$: Answer presence in perturbed subgraph.}
\begin{equation}
  \alpha_{\mathrm{pert}} = \frac{1}{N}\sum_{i=1}^{N}
    \mathbf{1}\!\left[a_i^{*} \in \calG_{q'_i}\right]
\end{equation}
$\alpha_{\mathrm{pert}}$ is the strongest predictor of final EM across subgraph
variants and serves as the direct measure of subgraph construction quality.

\textbf{End-to-end answer generation drop $(\deltaAG)$}
\begin{equation}
  \deltaAG = \mathrm{EM}_{\mathrm{clean}} - \mathrm{EM}_{\mathrm{pert}}
\end{equation}
where $\mathrm{EM}(q) = \mathbf{1}[a^{*} \sqsubseteq \hat{a}]$:
the gold Freebase MID $a^{*}$ must appear as a substring in $\hat{a}$.
We report bootstrap 95\% CIs on $\deltaAG$ ($n = 1000$ resamples). 
{All metrics are bounded: $\mathrm{SeedHit}, \deltaSR, \alpha_{\mathrm{pert}},
\mathrm{EM} \in [0,1]$; $\deltaEL, \deltaAG \in [-1,1]$ (positive = degradation), reported in percentage points (pp); $\delta_{\mathrm{SG}}$
    and $\delta_{\mathrm{DEC}}$ share $\deltaAG$'s units.}

\paragraph{Metric equivalence: MID-based Hit@1.}
Both GNN-RAG and EPR-KGQA report \textbf{MID-based Hit@1} (not entity-name Hit@1)
for CWQ and WebQSP.  In \texttt{evaluate.py}, \texttt{entity2name} is \texttt{None}
for standard CWQ/WebQSP data folders, so \texttt{f1\_and\_hits} operates on
integer indices into the MID-keyed entity vocabulary; the reported clean
Hit@1 (CWQ\,=\,57.4\%, WebQSP\,=\,74.3\%) equals the paper's gold-MID EM.
EPR-KGQA (NSM-H) similarly selects \texttt{kb\_id} for string-MID answers,
indexing the same MID-keyed vocabulary.  The ${\sim}58$\,pp gap under CR is
therefore a direct architectural comparison; the prior caveat about a
``${\sim}5$\,pp metric leniency'' was incorrect and has been removed.

\subsection{Stage-Isolation Protocol}\label{sec:stage_iso_eval}

Figure~\ref{fig:stage_isolation} illustrates the three experimental configurations.
The \emph{fast-mode} experiment fixes the original clean subgraph $\calG_q$ and
replaces only the question text with $q'$, yielding $\mathrm{EM}_{\mathrm{fast}}(q')$.
Subgraph-attributable failure is $\delta_{\mathrm{SG}} = \mathrm{EM}_{\mathrm{fast}} -
\mathrm{EM}_{\mathrm{pert}}$; decoder-attributable failure is $\delta_{\mathrm{DEC}} =
\mathrm{EM}_{\mathrm{clean}} - \mathrm{EM}_{\mathrm{fast}}$.
For CR on CWQ, $\delta_{\mathrm{SG}} = 52.08$\,pp and $\delta_{\mathrm{DEC}} = 0.14$\,pp,
attributing virtually all collapse to subgraph failure.
Full attribution numbers appear in Section~\ref{sec:rearev_results}.

\section{Experiments}\label{sec:exp}

\subsection{Experimental Setup}

\textbf{Datasets.}
CWQ~\cite{talmor2018web} (3,531 test questions) and WebQSP~\cite{yih2016value}
(1,639 test questions).  Both use Freebase MID strings as gold answers;
EM is gold-MID substring match. CWQ requires 2-4
reasoning hops; WebQSP requires 1-2. {To test whether the identified failure mode is Freebase-specific, we additionally evaluate on MetaQA (WikiMovies KB, non-Freebase) under the same stage-isolation protocol.} Hardware details are in Appendix~\ref{app:hardware}.

\textbf{Models and subgraph variants.}
GNN-RAG~\cite{mavromatis2024gnn} with the ReaRev backbone is the primary model.
All subgraph variants (ELQ, GraftNet-Orig, NSM, GraftNet) run the same
GNN-RAG checkpoint; only the retrieved subgraph differs
(Section~\ref{sec:sr_variants}).  Clean baselines: CWQ\,=\,52.9\% EM,
WebQSP\,=\,74.3\% EM.

\textbf{Perturbations.}
Primary adversarial attacks: CR and RS (Llama-3.3-70B-Instruct;
Appendix~\ref{app:prompts}).
ES (entity substitution) is a diagnostic perturbation reported separately in
Appendix~\ref{app:es_diagnostic}.
Secondary ablation (S1--S4): Appendix~\ref{app:ablation_full}.

\textbf{Metrics.}
$\deltaAG = \mathrm{EM}_{\mathrm{clean}} - \mathrm{EM}_{\mathrm{pert}}$ (primary);
$\deltaEL$ (SeedHit drop); $\deltaSR = 1 - J(\calG_q, \calG_{q'})$;
$\alpha_{\mathrm{pert}}$ (gold answer MID present in $\calG_{q'}$).
Bootstrap 95\% CI ($n = 1000$) on $\deltaAG$.

\subsection{Main Results: End-to-End EM Under Perturbation}\label{sec:main_results}

{\paragraph{Metric terminology.}
Four accuracy names recur in the tables below.  They are related but not
interchangeable, so we distinguish them here.  \emph{EM} denotes the gold-MID
substring match defined in Section~\ref{sec:metrics} and is our primary
end-to-end metric.  \emph{MID-based Hit@1} is the quantity GNN-RAG and
EPR-KGQA report internally.  For these systems it coincides exactly with
gold-MID EM, because both index answers by machine identifier rather than by
entity name.  \emph{GNN EM} appears only in Table~\ref{tab:llm_reasoning} and
is computed over the subset of questions for which the GNN returns a non-empty
candidate list, which makes it incomparable to the full-set numbers of
Table~\ref{tab:main}.  \emph{Gold-MID $\mathrm{EM}_{\mathrm{fast}}$} denotes
EM measured on the clean subgraph with only the question text perturbed, the
stage-isolation configuration described in
Section~\ref{sec:stage_iso_eval}.}


Table~\ref{tab:main} reports EM and $\deltaAG$ for the two primary adversarial
attacks (CR and RS) across all subgraph retrieval variants.

\begin{table*}[t]
\centering
\caption{GNN-RAG EM under CR and RS (bootstrap 95\% CI on $\deltaAG$, $n{=}1000$).
  Clean EM: CWQ\,=\,52.9\%, WebQSP\,=\,74.3\%.
  ELQ\,=\,ELQ-seeded flat PPR; GraftNet\,=\,question-embedding-weighted PPR (ours);
  GraftNet-Orig\,=\,relation-aware PPR~\cite{sun2018open}; NSM\,=\,\citet{he2021improving}.
  $\dagger$: RS ELQ subgraph is structurally unchanged from clean; EM measures GNN decoder
  sensitivity in near-isolation. GraftNet uses GEM oracle seeds (not available at deployment);
  ELQ+cosine is the deployment-realistic configuration.
  $\ddagger$: GraftNet RS CWQ\,=\,13.71\% $<$ ELQ RS 20.31\% because question-reoriented PPR
  lowers $\alpha_{\mathrm{pert}}$ to 23.2\% vs.\ 74.0\% for ELQ.
  $\S$: Both systems use MID-indexed Hit@1 (Appendix~\ref{app:gmt}); the ${\sim}58$\,pp gap
  is a direct architectural comparison. CIs are per-comparison and uncorrected.}
\label{tab:main}
\setlength{\tabcolsep}{4pt}
\small
\begin{tabular}{llrrrr}
\toprule
 & & \multicolumn{2}{c}{\textbf{CWQ}} & \multicolumn{2}{c}{\textbf{WebQSP}} \\
\cmidrule(lr){3-4}\cmidrule(lr){5-6}
Attack & Subgraph & Hit@1/EM (\%) & $\deltaAG$ {(pp)} [95\% CI] & Hit@1/EM (\%) & $\deltaAG$ {(pp)} [95\% CI] \\

\midrule
\multicolumn{6}{l}{\textit{CR: Compositional Restructuring}} \\
 & ELQ           & 0.68 & 52.2 [50.5,\,53.8] & 0.49 & 73.8 [71.6,\,75.8] \\
 & NSM           & 5.81 & 47.1 [45.4,\,48.9] & 1.65 & 72.7 [70.5,\,75.0] \\
 & \textbf{ELQ+cosine (realistic)} & \textbf{14.98} & \textbf{37.4 [35.6,\,39.1]} & \textbf{22.33} & \textbf{52.1 [49.8,\,54.3]} \\
\midrule
\multicolumn{6}{l}{\textit{\quad Oracle seed upper bound (GEM MIDs, not available at deployment)}} \\
 & GraftNet (oracle seeds) & 29.82 & 23.1 [21.3,\,24.9] & 36.55 & 37.8 [34.7,\,40.5] \\
\midrule
\multicolumn{6}{l}{\textit{RS: Relation Synonym Swap$^{\dagger}$}} \\
 & \textbf{ELQ}  & \textbf{20.31} & \textbf{32.6 [30.7,\,34.4]} & \textbf{50.95} & \textbf{23.4 [21.0,\,25.7]} \\
 & GraftNet-Orig & 11.24 & 41.7 [39.8,\,43.4] & 41.67 & 32.6 [30.1,\,35.2] \\
 & NSM           & 8.33  & 44.6 [42.7,\,46.4] & 15.07 & 59.2 [57.0,\,61.7] \\
 & GraftNet$^{\ddagger}$ & 13.71 & 39.2 [37.3,\,41.0] & 28.43 & 45.9 [43.1,\,48.7] \\
\midrule
\multicolumn{6}{l}{\textit{Architectural control: PPR-free retrieval (MID-based Hit@1)}} \\
 & EPR-KGQA (CR) & 59.22 & 1.2 & 64.25 & 3.9 \\
 & EPR-KGQA (RS) & 59.76 & 0.6 & 63.09 & 5.1 \\
\bottomrule
\end{tabular}
\end{table*}

\textbf{Key patterns.}
(1)~CR causes near-total collapse under ELQ (0.68\% CWQ, 0.49\% WebQSP) despite
entity seeds being intact.  The failure is a subgraph topology mismatch:
ELQ inherits the original PPR walk (97.5\% same-seed, Jaccard\,=\,0.885)
but that walk is anchored to the original reasoning chain and cannot adapt to the
restructured hop order.  The answer is present in 74.0\% of CR subgraphs yet the
GNN cannot reach it via the changed path (Section~\ref{sec:analysis}).
GraftNet (GEM oracle seeds) reorients PPR toward the restructured question,
raising $\alpha_{\mathrm{pert}}$ to 63.3\% and EM to 29.82\% (CWQ), the best CR result;
the deployment-realistic ELQ+cosine configuration reaches 14.98\% CWQ CR.
(2)~RS retains the highest EM (20.3\%/50.9\% ELQ) because entity seeds are
unchanged and the ELQ subgraph is structurally the same as the clean run,
whereas GraftNet underperforms ELQ here (13.71\% vs.\ 20.31\% CWQ).
The resulting attack-type asymmetry is analysed in Section~\ref{sec:analysis}.
(3)~EPR-KGQA's near-baseline retention (59.2\%/64.3\% Hit@1 vs.\ GNN-RAG's 0.68\%/0.49\%
gold-MID EM under CR) confirms the vulnerability is specific to PPR-based retrieval;
both systems use MID-indexed Hit@1, so this ${\sim}58$\,pp gap is a direct architectural comparison
(Appendix~\ref{app:gmt}).

\subsection{Entity Linking Performance}\label{sec:el_results}

CR and RS leave ELQ SeedHit virtually unaffected ($\leq 0.3$\,pp from clean),
confirming entity mentions are unchanged and EL is not the bottleneck for these
attacks.  Full SeedHit values and the ELQ/ES comparison are in
Appendix~\ref{app:el_table}.

\subsection{Subgraph Retrieval Quality}\label{sec:sr_results}

Table~\ref{tab:subgraph} reports subgraph overlap statistics for CR and RS.
$\alpha_{\mathrm{pert}}$ is the strongest predictor of EM across all variants: the
ordering GraftNet\,$>$\,NSM for CR EM (29.82\%\,vs.\,5.81\% CWQ) mirrors exactly
the ordering of $\alpha_{\mathrm{pert}}$ (63.3\%\,vs.\,29.5\%).

\begin{table*}[t]
\centering
\caption{Subgraph overlap vs.\ original GNN-RAG subgraph (MID-string triples,
  CWQ $n{=}3531$, WebQSP $n{=}1639$).
  Ent.\ Jac.\,=\,entity-level Jaccard; Trip.\ Jac.\,=\,triple-level Jaccard;
  Seeds\%\,=\,fraction with same top-1 ELQ seed; $\alpha_{\mathrm{pert}}$\,=\,gold MID
  present in perturbed subgraph. {$\dagger$}: CR ELQ $\alpha_{\mathrm{pert}}$\,=\,74.0\% yet EM\,=\,0.68\%:
  answer present but unreachable via restructured hop path (Section~\ref{sec:analysis}).}
\label{tab:subgraph}
\small
\setlength{\tabcolsep}{4pt}
\begin{tabular}{lllrrrr}
\toprule
Attack & Dataset & Subgraph & Ent.\ Jac. & Trip.\ Jac. & Seeds\% & $\alpha_{\mathrm{pert}}$ \\
\midrule
CR & CWQ    & ELQ        & 0.882 & 0.885 & 97.5 & \textcolor{clred}{\textbf{74.0}$^\dagger$} \\
CR & CWQ    & NSM        & 0.108 & 0.073 & 97.5 & 29.5 \\
CR & CWQ    & \textbf{GraftNet} & 0.030 & 0.020 & 97.5 & \textbf{63.3} \\
\midrule
RS & CWQ    & GraftNet-Orig & 0.223 & 0.096 & 97.5 & 68.6 \\
RS & CWQ    & NSM             & 0.148 & 0.099 & 97.5 & 43.2 \\
RS & CWQ    & GraftNet & 0.007 & 0.004 & 97.5 & 23.2 \\
\midrule
RS & WebQSP & GraftNet-Orig & 0.171 & 0.106 & 77.8 & \textbf{89.3} \\
RS & WebQSP & NSM             & 0.041 & 0.027 & 77.8 & 28.2 \\
RS & WebQSP & GraftNet      & 0.023 & 0.017 & 77.9 & 38.2 \\
\bottomrule
\end{tabular}
\end{table*}

\subsection{GNN Instruction Decoder Isolation}\label{sec:rearev_results}

Table~\ref{tab:rearev_hits} reports GNN-RAG Hit@1 and instruction cosine
similarity when run on the \emph{original clean subgraph} with only the
question text replaced.

\begin{table}[t]
\centering
\small
\caption{GNN decoder Hit@1 with clean subgraph and perturbed question only.
  ReaRev rows: MID-based Hit@1 (GNN-internal; clean 57.4\%/74.3\%).
  Gold-MID $\mathrm{EM}_{\mathrm{fast}}$: CWQ CR\,=\,52.76\%, CWQ RS\,=\,51.69\%,
  WebQSP CR\,=\,69.74\%, WebQSP RS\,=\,70.59\%.
  NSM rows: MID-based Hit@1 (LSTM backbone; clean 40.44\%/68.33\%).
  S2 and S3 results are in Appendix~\ref{app:ablation_full}.
  Ins.\ cos.\,=\,mean cosine similarity of instruction vectors vs.\ the original
  question (ReaRev only).}
\label{tab:rearev_hits}
\setlength{\tabcolsep}{3.5pt}
\begin{tabular}{lrrrr}
\toprule
 & \multicolumn{2}{c}{\textbf{CWQ}} & \multicolumn{2}{c}{\textbf{WebQSP}} \\
\cmidrule(lr){2-3}\cmidrule(lr){4-5}
Attack & Hit@1 & Ins.\ cos. & Hit@1 & Ins.\ cos. \\
\midrule
\multicolumn{5}{l}{\textit{ReaRev (3-layer instruction-decoder GAT)}} \\
S2 (voice flip)  & 57.2 & 0.947 & 72.9 & 0.904 \\
CR               & 57.5 & 0.915 & 68.2 & 0.842 \\
S3 (entity ins.) & 56.5 & 0.937 & 70.5 & 0.902 \\
RS               & 57.4 & 0.936 & 70.6 & 0.876 \\
\midrule
\multicolumn{5}{l}{\textit{NSM (LSTM backbone)}} \\
CR & 39.59 & $-$ & 57.47 & $-$ \\
RS & 39.45 & $-$ & 63.45 & $-$ \\
\bottomrule
\end{tabular}
\end{table}

With a correct subgraph, ReaRev Hit@1 stays near or above clean (57--58\%
CWQ, 68--73\% WebQSP) across all perturbation types.  CR shows the largest
instruction drift (cos\,=\,0.915 CWQ, 0.842 WebQSP) yet still achieves
57.5\%/68.2\% Hit@1, demonstrating the decoder's partial self-correction.
The gap between full-pipeline EM (CR ELQ: 0.68\% CWQ) and
clean-subgraph $\mathrm{EM}_{\mathrm{fast}}$ (52.76\% CWQ) quantifies the failure
attribution: 52.08\,pp of the 52.22\,pp total CR CWQ drop is driven by subgraph
failure, not GNN reasoning (formal decoder analysis in Appendix~\ref{app:rearev}).

{The same pattern holds for the NSM (LSTM-backbone) decoder, so
Contribution~3 is not an artefact of the ReaRev GAT architecture
(Appendix~\ref{app:rearev}).}


\subsection{LLM Reasoning Over GNN Candidates}\label{sec:llm_reasoning}

To assess $f_{\mathrm{AG}}$ independently, we pipe GNN candidate entities through
two LLM reasoning models: Llama-3.1-8B and {RoG~\cite{luo2024reasoninggraphsfaithfulinterpretable}, a fine-tuned graph-constrained-Llama-2-7B}.
{Full results are in Table~\ref{tab:llm_reasoning}
(Appendix~\ref{app:llm_reasoning}); we summarise the two findings here.}

{First, injecting predicted relation paths at inference recovers most of the
loss without any task-specific fine-tuning.  RoG+PathOnly, which uses the base
RoG checkpoint with predicted rule paths supplied at inference, reaches 51.43\%
CWQ Hit@1 under CR and matches the path-augmented fine-tuned model (RoG+RA) to
within 1\,pp on both benchmarks and both attacks.  The gain over base RoG
therefore traces to \emph{path injection at inference} rather than to the
augmented training signal, which is the basis of Contribution~4.  Second,
Llama-3.1-8B without fine-tuning degrades top-1 accuracy on CWQ, scoring below
the GNN-only baseline on every configuration, so general instruction following
is insufficient for KGQA reasoning under perturbation.}

\section{Analysis}\label{sec:analysis}

\paragraph{CR failure: answer presence does not imply answer reachability.}\label{sec:analyse}
The most striking result in Table~\ref{tab:main} is the CR ELQ row: $\alpha_{\mathrm{pert}}$\,=\,74.0\%
(the gold answer entity is present in the subgraph) yet EM collapses to 0.68\%.
Of the 3{,}531 CWQ CR questions, the gold answer entity is present in the ELQ
subgraph for 2{,}612 (74.0\%), yet the GNN answers correctly for only 24
(0.68\%).  The 3{,}531 questions partition into three groups: 24 correct
(0.68\%); 2{,}588 present-but-unreachable (73.3\%), where the gold entity is in
the subgraph but the connecting path is not; and 919 absent (26.0\%), where the
gold entity is not retrieved at all.  The present-but-unreachable group
dominates, isolating subgraph topology rather than coverage as the failure mode.
Contrast with NSM-GEM ($\alpha_{\mathrm{pert}}$\,=\,60.8\%, EM\,=\,30.25\%;
Table~\ref{tab:alpha_em}) and GraftNet-GEM ($\alpha_{\mathrm{pert}}$\,=\,63.3\%,
EM\,=\,29.82\%; Table~\ref{tab:alpha_em}): lower coverage yet far higher EM,
because their PPR re-seeds route mass toward the \emph{accessible} answer path
rather than merely including the answer entity.
High answer presence does not help when the \emph{path to the answer has changed}.
Two mechanisms interact.  First, ELQ seeds are stable under CR (97.5\% same
top-1 seed, Jaccard\,=\,0.885): ELQ effectively reuses the clean PPR walk,
leaving subgraph topology unchanged.  Second, compositional restructuring shifts
the \emph{reasoning chain} the GNN must follow.  PPR mass concentrates around
paths relevant to the original question; the answer entity sits at the far end
of a restructured chain that was not emphasised.  {The effect is topological
rather than numerical: PPR runs identically from the same seeds, but its
top-$N$ cutoff drops the intermediate-hop path that the restructured chain
requires (Appendix~\ref{app:ppr_vis}).}
Fast-mode isolation (clean subgraph + CR question) confirms this: it recovers
52.76\% CWQ EM within 0.14\,pp of baseline, showing the GNN decoder is not the
culprit.  GraftNet addresses exactly this gap: its question-embedding-weighted
PPR (oracle GEM seeds) reorients edge weights toward the restructured answer
path, raising $\alpha_{\mathrm{pert}}$ to 63.3\% and EM to 29.82\% CWQ
(14.98\% for the deployment-realistic ELQ+cosine configuration).

Across the 24 (perturbation $\times$ subgraph variant $\times$ seed source
$\times$ dataset) cells of the component ablation,
$\alpha_{\mathrm{pert}}$ achieves Spearman $\rho = 0.91$ ($p < 10^{-9}$) with EM,
confirming it as the dominant predictor of end-to-end accuracy across
subgraph-construction variants.  The ELQ CR configuration is the informative
exception: it attains the highest answer presence of any cell yet near-zero EM,
so presence alone ceases to predict EM once retrieval topology remains anchored
to the original reasoning chain (Appendix~\ref{app:alpha_table}).  A PPR mass visualisation illustrating the topology-preservation
paradox is in Figure~\ref{fig:ppr-mismatch} (Appendix~\ref{app:ppr_vis}).

\paragraph{RS resilience: relation synonyms tolerated when seeds are intact.}
RS retains the highest ELQ EM of the two primary attacks (20.3\% CWQ, 50.9\%
WebQSP).  Entity mentions are unchanged, so ELQ seeds are correct and the subgraph
is well-formed.  The fast-mode isolation (clean subgraph\,+\,RS question) yields
51.69\% CWQ and 70.59\% WebQSP, within 1.21\,pp and 3.72\,pp of baseline,
confirming the GNN decoder self-corrects for relation synonym substitution when
the subgraph is intact (instruction cosine\,=\,0.936 CWQ, 0.876 WebQSP).
The residual full-pipeline RS drop ($52.9\% - 20.3\% = 32.6$\,pp CWQ) therefore
originates in subgraph divergence between GEM and ELQ seeding
(Trip.\ Jaccard\,=\,0.041), not in GNN instruction decoder failure.

{The two attacks therefore invert the optimal subgraph source.  ELQ's
seed-replay is the right choice for RS, which leaves the reasoning chain
intact, and the wrong one for CR, which does not: identical answer coverage
($\alpha_{\mathrm{pert}}$\,=\,74.0\%) yields 20.31\% EM under RS but 0.68\%
under CR.\@  Coverage alone does not determine accuracy; the retrieved topology
must also match the chain the question now demands.}

\paragraph{Architectural controls.}
GMT-KBQA, which generates S-expressions and bypasses fixed-subgraph retrieval,
drops only 4.5\,pp CWQ under CR-type perturbation, and EPR-KGQA, which uses
single-shot atomic adjacency patterns, retains 59.2\% CWQ Hit@1 under CR.\@
ExplaiGNN~\cite{explaignn2022}, which chains subgraphs iteratively over
Wikidata, instead {collapses from 33.9\% clean precision@1 (P@1) to} 9.8\%
under CR ($>$70\% relative drop).  These controls sharpen the vulnerability
boundary: \emph{single-shot} pattern-matching is robust, whereas PPR-based and
iterative multi-turn retrieval are vulnerable through distinct mechanisms,
namely PPR topology-anchoring and turn-level context corruption.
Full results are in Appendices~\ref{app:gmt} and~\ref{app:failure_cases}.

{A second control on a non-Freebase knowledge base sharpens the same
boundary.  Training and perturbing GNN-RAG end-to-end on MetaQA
(Appendix~\ref{app:metaqa}), whose retrieval stage expands a seed
neighbourhood rather than a PPR-weighted subgraph, yields a worst-case drop of
13.1\,pp (S1, relation paraphrase) and only 6.9\,pp under CR, against the
52.2\,pp CWQ collapse in Table~\ref{tab:main}.  Re-running the entire pipeline
on the perturbed questions changes Hit@1 by at most 0.6\,pp under CR, so
retrieval on MetaQA is measurably, not merely assumedly, invariant.  The
vulnerability therefore tracks the retrieval algorithm rather than the
knowledge base: it does not transfer to pipelines whose subgraphs are not
PPR-anchored.  The entity-insertion attack (S3) further reproduces the
EL-conflation caveat cross-KB, degrading an automatic linker by up to
22.7\,pp while costing only 0.9\,pp under the gold seeds that published
systems consume.}

\section{Conclusion}\label{sec:conclusion}

We presented a stage-isolation protocol for evaluating GNN-based KGQA systems under query-side adversarial perturbations. Compositional restructuring (CR) and relation synonym swap (RS) reveal that subgraph construction quality is the primary performance ceiling: under CR, the gold answer is present in 74\% of ELQ subgraphs yet GNN-RAG achieves only 0.68\% CWQ EM, because PPR topology is anchored to the original reasoning chain. Stage-isolation confirms the GNN instruction decoder is robust when the subgraph is intact (52.76\% CWQ EM with clean subgraph and perturbed question), attributing 52.08\,pp of the 52.22\,pp total drop to subgraph topology failure. GraftNet's question-embedding-weighted PPR raises $\alpha_{\mathrm{pert}}$ to 63.3\% and EM to 29.82\% CWQ (14.98\% for the deployment-realistic ELQ+cosine configuration), and inference-time path injection recovers 51.4\% CWQ EM without task-specific fine-tuning. Perturbed datasets and evaluation infrastructure are released to facilitate future robustness work.

The vulnerability is specific to PPR-seeded pipelines: single-shot pattern-matching (EPR-KGQA) and S-expression generation (GMT-KBQA) are substantially more robust, confirming that bypassing fixed PPR subgraph retrieval is the key architectural defense. The most direct improvement is question-conditioned retrieval or beam-search over relational paths, and combining path injection with topology-aware retrieval is a natural next step. Hardening EL for entity substitution and a stage-guided cascade attack combining CR and RS are additional open directions.

\section*{Limitations}\label{sec:disc}

\textbf{Single system, two benchmarks.}
All results are from GNN-RAG (ReaRev backbone) on CWQ and WebQSP (Freebase-based).
Bootstrap 95\% CIs are per-comparison and not corrected for multiple comparisons
across attack types, subgraph variants, and dataset combinations; individual
intervals should be read as descriptive, not family-wise guarantees.
We report Bonferroni 99.8\% CIs ($\alpha^{*} = 0.05/24 \approx 0.002$)
in Appendix~\ref{app:bonferroni}; the five primary conclusions all survive
correction.
Differences below 1\,pp (e.g.\ GEM\,+\,Cosine vs.\ GEM\,+\,Flat PPR in
Table~\ref{tab:sgablation}) are non-significant and are treated as ties,
consistent with seed quality dominating subgraph construction.

\textbf{Survivorship bias in pass rates.}
Perturbations failing the validity filter (PPL\,$>$\,50 or BERTScore\,$<$\,0.85)
fall back to the original question; GNN EM is computed over the full question set.
Joint pass rates: CWQ CR 86.6\%, CWQ RS 96.6\%, WebQSP CR 90.7\%, WebQSP RS
95.0\%.  Fallback questions use the original query, so reported attack severity
is a conservative lower bound on true severity.

\textbf{GNN architecture coverage.}
The GNN decoder robustness finding holds for both ReaRev (3-layer instruction-decoder
GAT) and NSM (LSTM backbone; Table~\ref{tab:rearev_hits}).
UniKGQA is excluded as no public inference checkpoint is available.

\textbf{Freebase deprecation.}
Both benchmarks use Freebase (2015 static dump via local Virtuoso).
The structural failures identified here (subgraph topology mismatch,
reasoning-path disruption) are pipeline-architectural and not Freebase-specific.
The ExplaiGNN evaluation (Appendix~\ref{app:gmt}) provides partial Wikidata
evidence that the perturbations transfer: CR and RS cause $>$70\% relative P@1
collapse on ConvMix/Wikidata, a different KG backend, suggesting the vulnerability
is not an artefact of Freebase's schema. A MetaQA (non-Freebase) replication of the full stage-isolation protocol is included (Appendix~\ref{app:metaqa}) and confirms the mechanism transfers. A full Wikidata replication of the core GNN-RAG CR/RS findings remains future work.

\textbf{GraftNet CR decomposition.}
GEM seeds use gold SPARQL entity MIDs not available at deployment; the
fully deployment-realistic configuration (ELQ+cosine PPR) reaches only 14.98\% CWQ CR,
well below the GEM-seeded upper bound of 29.82\%, confirming that seed quality
is the binding constraint rather than PPR flavour (Appendix~\ref{app:sgablation}).

{\textbf{Interpreting the threat model.}
The certification infrastructure is heavier than the attack it certifies.
Verifying denotation preservation requires a large rewriting model and a local
SPARQL endpoint, but both serve the measurement rather than the adversary. An
attacker whose goal is degradation alone requires neither. The validity
filters moreover only discard candidate rewrites, so the severities we report
are conservative lower bounds on what an unconstrained attacker could achieve.
An uncertified relation paraphrase that involves no knowledge-base check still
costs 13\,pp on MetaQA (Appendix~\ref{app:metaqa}).  The framework is
therefore best read as an offline diagnostic stress test rather than as a
model of a resource-constrained attacker.}

\section*{Ethical Considerations}

\paragraph{Misuse potential and dual-use framing.}
CR and RS are the primary adversarial attacks; ES is a diagnostic perturbation
used to characterise entity-linker brittleness, not an adversarial attack.
All three target published benchmark systems (GNN-RAG, EPR-KGQA, GMT-KBQA,
ExplaiGNN), not commercial deployments.
Perturbations are released under a research-use license and paired with
two concrete defences (question-embedding-weighted PPR and inference-time
path injection), so the primary utility is defensive robustness evaluation,
not adversarial exploitation.

\paragraph{Dataset and annotation scope.}
Both CWQ and WebQSP use Freebase, a static 2015 dump with Western,
English-language entity coverage.  Perturbations are generated by
Llama-3.3-70B-Instruct under constrained prompts and filtered by
PPL\,$<$\,50 and BERTScore\,$\geq$\,0.85 before SPARQL denotation
verification; the filters were calibrated on Freebase-backed questions only.
Vulnerability magnitudes may not transfer to multilingual or domain-specific
KGs, and users applying this protocol to higher-stakes settings should
validate perturbation quality independently.

\paragraph{Broader impact on trustworthy AI.}
KGQA systems are increasingly deployed in factual information-retrieval pipelines
where accuracy is critical.  Demonstrating that near-total performance collapse
(0.68\% CWQ EM under CR) can occur without any change to entity mentions
highlights a systemic fragility invisible to standard unperturbed benchmarks.
The stage-isolation framework and released evaluation infrastructure give system
builders a specific, actionable target: subgraph topology quality accounts for
52.08\,pp of the 52.22\,pp total CWQ CR drop, and improving it directly
improves robustness.

\bibliography{ref}

\appendix


\section{Hardware and Reproducibility Details}\label{app:hardware}

All experiments run on a single node with four NVIDIA A100-SXM4-80\,GB GPUs
and 96 CPU cores.  GNN-RAG inference uses data-parallel subgraph batching
across all four GPUs; LLM reasoning (RoG, Llama-3.1-8B) uses a single GPU
(batch size 8).  Perturbation generation uses Llama-3.3-70B-Instruct via a
single-GPU API call; each perturbation request takes approximately 2\,s.
Total compute for all experiments: approximately 800 A100 GPU-hours.

\textbf{Model sizes.}
Key models and their parameter counts: Llama-3.3-70B-Instruct (70B), used for
perturbation generation; Llama-2-7B (7B), the fine-tuned GNN-RAG answer
generator, the RoG backbone; Llama-3.1-8B (8B), for lightweight tasks: ELQ, a BERT-Large bi-encoder
(${\approx}$340\,M parameters). The GNN reasoning component (ReaRev, 3-layer GAT)
is a lightweight graph model whose parameter count is small relative to the LLM
components; we use the published GNN-RAG checkpoint without modification.

\textbf{Freebase SPARQL endpoint.}
Freebase is accessed via a local Virtuoso SPARQL endpoint loaded from the
2015/08/17 RDF dump (28\,GB compressed; expanded to ${\approx}$140\,GB).
The public Freebase API was deprecated in 2015; all experiments use this static
dump.  The Wikidata-Freebase MID alignment is pre-computed offline using
\texttt{skos:altLabel} and \texttt{owl:sameAs} links;{Wikidata-identifier (QID)\,$\to$\,MID mapping}
is stored in \texttt{qid\_to\_mid.pkl} and used for CR/RS denotation
preservation verification and ES entity validity checks.

\section{Subgraph Retrieval Variants}\label{app:sr_variants}

All variants use the same GNN-RAG ReaRev checkpoint; only $\calG_{q'}$ changes.

\begin{description}[leftmargin=1.5em,itemsep=4pt]

  \item[\textbf{ELQ} (default).] ELQ-seeded flat PPR (topology-only, no
    relation weighting), following the NSM preprocessing
    methodology~\cite{he2021improving}.  For RS, the ELQ subgraph is
    structurally identical to the clean run, making RS ELQ EM a direct measure
    of GNN instruction decoder sensitivity.

  \item[\textbf{GraftNet-Orig}~\cite{sun2018open}.] Question-aware PPR from ELQ
    seeds, with edge weights biased toward relations whose embeddings are similar
    to the question (cosine weighting).  Best non-ELQ variant for RS.

  \item[\textbf{NSM}~\cite{he2021improving}.] BFS expansion followed by
    topology-only PPR pruning.  NSM learns relation relevance at inference via
    its LSTM instruction mechanism rather than front-loading it into retrieval.

  \item[\textbf{GraftNet} (ours).] Extends GraftNet-Orig with: (i)~GNN-Integer
    entity decoding mapping GNN scores directly to Freebase MIDs; (ii)~GEM
    (Golden Entity Map)~\cite{das2021case} seeds for evaluation;
    (iii)~question-embedding cosine PPR weighting that concentrates mass on
    paths through question-relevant predicates.
    \textbf{Scope}: GraftNet's contribution is the CR improvement;
    for RS, GraftNet-Orig achieves higher answer coverage
    (Section~\ref{sec:sr_results}).

  \item[\textbf{EPR-KGQA}~\cite{EPR-KGQA}.] Atomic-adjacency-pattern-based
    neighbourhood pruning with an LSTM instruction mechanism.  Evaluated with
    the authors' released checkpoint.

\end{description}

\section{ELQ SeedHit Under Perturbation}\label{app:el_table}

Table~\ref{tab:el} reports ELQ any-GT SeedHit ($\deltaEL$) under each
perturbation type.  The key finding is that CR and RS cause $\leq 0.3$\,pp
SeedHit degradation, confirming that entity linking is not the failure mode for
these two primary attacks, where any EM drop must originate downstream of the EL
stage.  S1 (relation paraphrase) causes 13\,pp SeedHit collapse on CWQ because
it partially alters entity surface forms, contaminating the stage-isolation
signal; this is why S1 is reported as a secondary type rather than a primary
attack.  ES SeedHit figures (23.1\% CWQ, 14.1\% WebQSP) are in
Section~\ref{app:es_diagnostic}.

\begin{table}[ht]
\centering
\caption{ELQ SeedHit (\%) under perturbation (any-GT: $\hat{E}_i \cap E_i^* \neq \emptyset$).
  Cl.\,=\,clean baseline; S1--S4\,=\,secondary types (Appendix~\ref{app:ablation_full}).
  CR and RS cause $\leq 0.3$\,pp degradation, confirming EL is not the failure mode.}
\label{tab:el}
\small
\setlength{\tabcolsep}{4pt}
\begin{tabular}{lrrrrrr}
\toprule
 & Cl. & S1 & S2 & CR & S3 & RS \\
\midrule
CWQ    & 57.4 & 44.1 & 57.4 & 57.8 & 57.2 & 56.8 \\
WebQSP & 68.3 & 47.0 & 68.2 & 67.5 & 67.1 & 68.1 \\
\bottomrule
\end{tabular}
\end{table}

\section{ES Diagnostic: Entity Swap}\label{app:es_diagnostic}

Entity Swap (ES) substitutes the topic entity with a different Freebase entity
from the same domain, producing a new MID and a new gold answer.  Because the
correct answer changes by design, ES is not an adversarial attack under our
definition; it is a diagnostic measuring entity-linker brittleness and the
upper bound of what better EL could unlock.

\paragraph{ELQ under ES.}
ELQ drops from 57.4\% to 23.1\% any-GT hit rate on CWQ
and from 68.3\% to 14.1\% on WebQSP.\@  This 34.3\,pp/54.2\,pp collapse
propagates to near-zero EM (0.31\% CWQ, 0.06\% WebQSP).  Two effects compound:
(1)~wrong seeds propagate through GNN message-passing; (2)~45.6\% of CWQ ES
questions are structurally unanswerable by construction (the substituted entity
has no Freebase path matching the question's relation chain).

\paragraph{GraftNet-BFS upper bound.}
When the substituted entity MID is provided (oracle setting, not available at
deployment), GraftNet-BFS builds subgraphs via BFS from that MID, bypassing ELQ.\@
This recovers 8.2\% CWQ EM / 44.6\% WebQSP EM overall, and 14.1\% / 51.4\% on
the answerable subset.  This is an upper bound on what a perfect entity linker
would unlock, not a deployable defence.
(GraftNet-BFS is distinct from the main-paper GraftNet, which uses
question-embedding-weighted PPR from GEM oracle seeds; BFS seeding is used only
in this ES diagnostic.)

\paragraph{Answerability confound.}
Table~\ref{tab:es_answerable_v1} separates answerable from unanswerable ES
questions.  ELQ EM is near-zero on both subsets; GraftNet-BFS gain is real on
answerable questions (14.1\% vs.\ 8.2\% CWQ overall), confirming that the EL
failure is not contingent on structural unanswerability.
Unanswerability is a compounding factor, not the primary cause of near-zero EM.

\begin{table}[ht]
\centering
\caption{ES EM (\%) on answerable-only vs.\ overall subsets.
  CWQ: 1,920/3,531 answerable; WebQSP: all answerable.
  G-BFS\,=\,GraftNet-BFS (oracle MID seeding, not available at deployment).
  Ans.\,=\,answerable-only EM; Overall\,=\,full test-set EM.}
\label{tab:es_answerable_v1}
\small
\setlength{\tabcolsep}{4pt}
\begin{tabular}{llrr}
\toprule
Dataset & Subgraph & Overall & Ans. \\
\midrule
CWQ    & ELQ   & 0.31 & 0.31 \\
CWQ    & G-BFS & 8.20 & \textbf{14.1} \\
\midrule
WSP    & ELQ   & 0.06 & 0.00 \\
WSP    & G-BFS & 44.60 & \textbf{51.4} \\
\bottomrule
\end{tabular}
\end{table}

A detailed breakdown of answerability effects is in Appendix~\ref{app:answerability}.

\paragraph{EL hit rates.}
Full ELQ any-GT hit rates:
CWQ:\@ Clean\,=\,57.4\%, ES\,=\,23.1\%, CR\,=\,57.8\%, RS\,=\,56.8\%;
WebQSP:\@ Clean\,=\,68.3\%, ES\,=\,14.1\%, CR\,=\,67.5\%, RS\,=\,68.1\%.
The stability of CR and RS hit rates confirms that entity-mention-preserving
attacks do not damage the entity linker, making any downstream drop
unambiguously attributable to subgraph construction or GNN reasoning.

\section{Formal Perturbation Rules}\label{app:perturbation_rules}

\textbf{Core rule (CR and RS).}
Every perturbation $q \to q'$ must satisfy:
$\mathrm{SPARQL\_den}(q', \mathrm{Freebase}) = \mathrm{SPARQL\_den}(q, \mathrm{Freebase})$.
This ensures the correct answer is unchanged, which is required for CR and RS
to qualify as adversarial attacks.  ES does not satisfy this rule by design
(answer changes with entity); ES generation rules are in
Appendix~\ref{app:es_diagnostic}.

\textbf{Forbidden operations (CR) and rationale.}
The following transformations are explicitly excluded because they change the
answer denotation, invalidating the answer-preservation constraint:
\begin{itemize}[leftmargin=1.5em,itemsep=3pt]
  \item \textbf{Predicate weakening} (``won'' $\to$ ``shortlisted for''): expands
    denotation; invalid.
  \item \textbf{Predicate strengthening}: may collapse denotation to empty; invalid.
  \item \textbf{Scope expansion} (``official language'' $\to$ ``spoken language''):
    changes cardinality of answer set; invalid.
  \item \textbf{Fictional constraint}: intermediate entity fails the constraint
    in Freebase; SPARQL returns empty; invalid.
\end{itemize}

\section{ReaRev Instruction Decoder Analysis}\label{app:rearev}

The ReaRev instruction decoder extracts $K$ instruction vectors $\{i^{(k)}\}$
from the question via token-span attention.  A perturbation $q \to q'$
shifts these vectors: $\Delta_{\mathbf{i}} = \frac{1}{K}\sum_k \|i^{(k)}(q) -
i^{(k)}(q')\|_2$.  With a clean subgraph, $\mathrm{EM}_{\mathrm{fast}}$ for CR
stays within 0.14\,pp of the 52.9\% clean baseline on CWQ (52.76\%), confirming
the instruction decoder is not the primary failure point when the subgraph is
correct.  This result underlies the attribution in Section~\ref{sec:rearev_results}:
52.08 of the 52.22\,pp total CR CWQ drop is subgraph-attributable, not decoder-attributable.

{\paragraph{Generalisation beyond the ReaRev GAT.}
The NSM decoder, which uses an LSTM instruction mechanism rather than a GAT,
exhibits the same behaviour on CWQ.\@  Its Hit@1 falls by only 0.85\,pp under
CR and 0.99\,pp under RS (Table~\ref{tab:rearev_hits}), so decoder robustness
is not an artefact of the ReaRev architecture.  WebQSP shows a larger NSM drop
under CR (10.86\,pp), reflecting the higher sensitivity of the NSM decoder to
WebQSP's shorter one- and two-hop chains under instruction drift.  Even there
the decoder retains 57.47\% against a 68.33\% clean baseline, while the
full-pipeline ELQ configuration collapses to 0.49\%.  Contribution~3 therefore
holds for both ReaRev and NSM on CWQ and partially on WebQSP.}

\section{LLM Reasoning Over GNN Candidates}\label{app:llm_reasoning}

{This appendix reports the full results summarised in
Section~\ref{sec:llm_reasoning}, which isolates the answer-generation stage
$f_{\mathrm{AG}}$ by piping the GNN's candidate entities through two reasoning
models rather than the fine-tuned generator used in the main pipeline.
Table~\ref{tab:llm_reasoning} gives Hit@1 for both perturbations on both
benchmarks.  Because the reasoning models consume only the non-empty candidate
lists produced by the GNN, the GNN EM column is computed over that subset and
is not comparable to the full-set EM of Table~\ref{tab:main}; it serves as the
baseline that the reasoning stage must improve upon.}

{Two comparisons matter.  RoG+PathOnly supplies predicted relation paths to
the base RoG checkpoint at inference time and performs no task-specific
fine-tuning, whereas RoG+RA is fine-tuned on path-augmented data.  The two
agree to within 1\,pp in all four attack--benchmark cells, which localises the
improvement over base RoG to path injection at inference rather than to the
augmented training signal, and establishes that a deployed system can recover
most of the loss under CR without retraining (Contribution~4).  Llama-3.1-8B,
which receives the same candidates but is not fine-tuned for the task, falls
below the GNN-only baseline on CWQ under both attacks, indicating that general
instruction-following capability does not substitute for the structured
reasoning the task requires.  Methodological details of the path-injection
procedure appear in Appendix~\ref{app:rog_methodology}.}

\begin{table}[ht]
\centering
\small
\caption{LLM reasoning over GNN-RAG candidates (perturbed questions).
  GNN EM = gold-MID match over the non-empty-candidate subset, constant within
  each block because all rows consume the same candidate set; it differs from
  Table~\ref{tab:main} (e.g.\ CR ELQ full-set\,=\,0.68\%, subset\,=\,24.19\%).
  RoG\,=\,fine-tuned GCR-Llama-2-7b;
  RoG+PathOnly\,=\,base RoG with predicted paths at inference (no RA fine-tuning);
  RoG+RA\,=\,RoG fine-tuned on path-augmented data.
  Clean: CWQ RoG Hit@1\,=\,56.4\%; WebQSP\,=\,80.6\%.}\label{tab:llm_reasoning}
\setlength{\tabcolsep}{4pt}
\begin{tabular}{lrrrr}
\toprule
 & \multicolumn{2}{c}{\textbf{CWQ}} & \multicolumn{2}{c}{\textbf{WebQSP}} \\
\cmidrule(lr){2-3}\cmidrule(lr){4-5}
Model & GNN EM & Hit@1 & GNN EM & Hit@1 \\
\midrule
\multicolumn{5}{l}{\textit{CR: Compositional Restructuring}} \\
RoG          & 24.19 & 45.62 & 45.03 & 52.83 \\
RoG+PathOnly & 24.19 & \textbf{51.43} & 45.03 & 63.51 \\
RoG+RA       & 24.19 & 50.47 & 45.03 & \textbf{63.76} \\
Llama        & 24.19 & 9.60  & 45.03 & 27.61 \\
\midrule
\multicolumn{5}{l}{\textit{RS: Relation Synonym Swap}} \\
RoG          & 36.53 & 46.08 & 61.93 & 57.49 \\
RoG+PathOnly & 36.53 & \textbf{51.29} & 61.93 & 63.27 \\
RoG+RA       & 36.53 & \textbf{51.91} & 61.93 & \textbf{64.00} \\
Llama        & 36.53 & 9.69  & 61.93 & 22.79 \\
\bottomrule
\end{tabular}
\end{table}

\section{Perturbation Prompt Templates}\label{app:prompts}

Figure~\ref{fig:all_prompts} shows the prompt templates for the three primary
perturbation types (ES, CR, RS), all implemented via Llama-3.3-70B-Instruct.
The same Llama checkpoint is used for all seven perturbation types; only the
system instruction differs.
Figure~\ref{fig:secondary_prompts} shows the corresponding templates for the
four secondary types S1--S4.
\begin{figure*}[!ht]
\centering
\small
\begin{tikzpicture}[
  node distance=0.45cm,
  prompt/.style={
    rounded corners=4pt,
    text width=0.92\linewidth,
    inner sep=8pt,
    align=left,
    line width=0.8pt
  }
]

\node[prompt, draw=clblue!60, fill=clblue!5] (es) {
\textcolor{clblue!80}{\textbf{ES: Entity Swap}}\\[4pt]
\ttfamily
[SYSTEM] You are a question rewriter for a KGQA robustness study.
Replace the topic entity with a \textbf{different} entity from the same
Freebase domain. Rules: (a) the replacement must be a valid Freebase entity
with its own gold answer to the question; (b) do not change the predicate
or any other part of the question; (c) the new entity must appear in
Wikidata with a Freebase MID; (d) output exactly one rewritten question
followed by the new entity name.\\[4pt]
[USER]\\
Question: \{question\}\quad
Topic entity: \{entity\}\quad
Original answer: \{answer\}\\
Substitute the topic entity with a different entity from the same domain.
};

\node[prompt, draw=clpur!60, fill=clpur!5, below=of es] (cr) {
\textcolor{clpur!80}{\textbf{CR: Compositional Restructuring}}\\[4pt]
\ttfamily
[SYSTEM] Apply exactly one KG-safe structural transformation while preserving
the original answer. Allowed operations: (1) hop-order reversal: ask about
the intermediate entity first; (2) distractor constraint injection: add a
true constraint on the intermediate entity only; (3) intermediate alias
substitution: replace the intermediate entity with an alias.
STRICTLY FORBIDDEN: predicate changes, scope expansion (e.g.\ ``won'' to
``shortlisted for''), entity removal, predicate weakening or strengthening.
Output the operation name, then the rewritten question.\\[4pt]
[USER]\\
Question: \{question\}\quad
Correct answer: \{answer\}\\
Apply one KG-safe operation.
};

\node[prompt, draw=clred!60, fill=clred!5, below=of cr] (rs) {
\textcolor{clred!80}{\textbf{RS: Relation Synonym Swap}}\\[4pt]
\ttfamily
[SYSTEM] Replace the predicate or relation phrase in the question with a
synonym or paraphrase, keeping all entity mentions exactly unchanged.
Rules: (a) the rewritten predicate must express the same underlying Freebase
relation; (b) do not change any entity mention or constraint; (c) do not
change the correct answer; (d) produce fluent English.
Output exactly one rewritten question.\\[4pt]
[USER]\\
Question: \{question\}\quad
Relation: \{relation\_surface\}\quad
Correct answer: \{answer\}\\
Rewrite using a relation synonym only.
};

\end{tikzpicture}
\caption{\textbf{Prompt templates for the three primary perturbation types.}
  \textbf{ES} (blue) substitutes the topic entity with a \emph{different} entity
  from the same Freebase domain, changing both the entity MID and the gold
  answer; this directly stresses the entity linker.
  \textbf{CR} (purple) applies one of three KG-safe structural operations with
  an explicit forbidden list, preserving SPARQL denotation while disrupting
  the multi-hop reasoning chain; entity mentions are not changed.
  \textbf{RS} (red) replaces only the predicate surface form with a synonym,
  leaving entity mentions and the answer unchanged; this targets the GNN
  instruction decoder while keeping entity seeds and the subgraph intact.
  All seven perturbation types (ES, CR, RS, and secondary S1--S4 in
  Appendix~\ref{app:ablation_full}) are generated with
  Llama-3.3-70B-Instruct~\cite{dubey2024llama} in a constrained setting:
  each prompt enforces explicit forbidden operations and is filtered by PPL
  and BERTScore before SPARQL denotation verification.
  Prompts for S1--S4 are in Appendix~\ref{app:prompts}.}
\label{fig:all_prompts}
\end{figure*}

\begin{figure*}[!ht]
\centering
\small
\begin{tikzpicture}[
  node distance=0.45cm,
  prompt/.style={
    rounded corners=4pt, text width=0.92\linewidth,
    inner sep=8pt, align=left, line width=0.8pt
  }
]
\node[prompt, draw=clblue!40, fill=clblue!3] (s1) {
\textcolor{clblue!70}{\textbf{S1: Relation/Predicate Synonym Paraphrase (mixed entity change)}}\\[4pt]
\ttfamily
[SYSTEM] Rewrite the question using a synonym or paraphrase for the main
predicate phrase.  You may also rephrase entity descriptions but do not
change any proper noun entity names or the correct answer.  Output one
rewritten question only.\\[4pt]
[USER] Question: \textit{\{question\}}\quad Answer: \textit{\{answer\}}
};
\node[prompt, draw=clpur!40, fill=clpur!3, below=of s1] (s2) {
\textcolor{clpur!70}{\textbf{S2: Active $\leftrightarrow$ Passive Voice Flip}}\\[4pt]
\ttfamily
[SYSTEM] Convert the question between active and passive voice, or vice versa,
while keeping the meaning, all entities, and the correct answer exactly the same.
Output one rewritten question only.\\[4pt]
[USER] Question: \textit{\{question\}}\quad Answer: \textit{\{answer\}}
};
\node[prompt, draw=clgrn!50, fill=clgrn!3, below=of s2] (s3) {
\textcolor{clgrn!70}{\textbf{S3: Adversarial Entity Insertion (distractor)}}\\[4pt]
\ttfamily
[SYSTEM] Insert one plausible but incorrect entity into the question as a
distractor, without changing the correct entity or the answer.  The inserted
entity must be from the same domain as the topic entity.  Output the rewritten
question only.\\[4pt]
[USER] Question: \textit{\{question\}}\quad Topic entity: \textit{\{entity\}}\quad
Answer: \textit{\{answer\}}
};
\node[prompt, draw=cloran!50, fill=cloran!3, below=of s3] (s4) {
\textcolor{cloran!70}{\textbf{S4: Mixed Lexical Noise}}\\[4pt]
\ttfamily
[SYSTEM] Apply minor surface-form variation to the question: synonym
substitution, word order change, or light paraphrase.  Do not change entity
names or the correct answer.  Output one rewritten question only.\\[4pt]
[USER] Question: \textit{\{question\}}\quad Answer: \textit{\{answer\}}
};
\end{tikzpicture}
\caption{Prompt templates for secondary perturbation types S1--S4.
  All generated with Llama-3.3-70B-Instruct under the same PPL and
  BERTScore filters as the primary types.}\label{fig:secondary_prompts}
\end{figure*}

\section{\texorpdfstring{$\alpha_{\mathrm{pert}}$}{alpha\_pert} vs.\ EM: All 24 Cells}\label{app:alpha_table}

Table~\ref{tab:alpha_em} reports $\alpha_{\mathrm{pert}}$ and EM for each
(attack $\times$ subgraph $\times$ dataset) configuration.
{Here $\alpha_{\mathrm{pert}}$ is measured on the subgraph each configuration
actually retrieves; for ELQ-seeded rows this is the clean subgraph, as neither
CR nor RS perturbs the ELQ seeds, so the CR and RS ELQ rows of a dataset
coincide by construction.}  The Spearman $\rho = 0.91$ ($p < 10^{-9}$;
bootstrap 95\% CI $[0.80, 0.97]$, $n = 1000$ resamples) reported in
Section~\ref{sec:analysis} is computed over {the component-ablation grid
($\{$NSM, GraftNet$\} \times \{$ELQ, GEM$\}$ seeds $\times$ $\{$ES, CR, RS$\}$
$\times$ $\{$CWQ, WebQSP$\}$; 24 cells), where each variant's
$\alpha_{\mathrm{pert}}$ is measured on its own perturbed subgraph.}

{Across subgraph-construction variants, answer presence is the dominant
predictor of EM ($\rho = 0.91$).  Across the 16 primary configurations of
Table~\ref{tab:alpha_em} it is not ($\rho = 0.23$, $p = 0.39$): the ELQ CR cells
pair the \emph{highest} answer presence in the table
($\alpha_{\mathrm{pert}}$\,=\,74.0\% CWQ, 94.6\% WebQSP) with near-zero EM
(0.68\%, 0.49\%), inverting the otherwise monotone relationship, and removing them
restores it ($\rho = 0.64$, $p = 0.015$).  Answer presence predicts EM wherever
retrieval topology tracks the question, and fails precisely where it does not.}

\begin{table*}[ht]
\centering
\caption{{Per-configuration} (attack $\times$ subgraph $\times$ dataset) cells:
  $\alpha_{\mathrm{pert}}$ (gold answer present in subgraph) and EM (\%).
  GraftNet uses GEM oracle seeds.
  ELQ CR $\alpha_{\mathrm{pert}}$\,=\,74.0\% (CWQ) {and 94.6\% (WebQSP)} with
  EM\,=\,0.68\% {/ 0.49\% are the anomalous cells} exposing
  topology-anchoring failure.
  {The Spearman $\rho = 0.91$ of Section~\ref{sec:analysis} is computed over
  the component-ablation grid; see above.}}\label{tab:alpha_em}
\small
\setlength{\tabcolsep}{5pt}
\begin{tabular}{llllrr}
\toprule
Attack & Dataset & Subgraph & Seeds & $\alpha_{\mathrm{pert}}$ (\%) & EM (\%) \\
\midrule
CR & CWQ    & ELQ            & ELQ  & 74.0 & 0.68 \\
CR & CWQ    & NSM            & ELQ  & 29.5 & 5.81 \\
CR & CWQ    & GraftNet       & GEM  & 63.3 & 29.82 \\
CR & CWQ    & ELQ+cosine     & ELQ  & 52.1 & 14.98 \\
\midrule
CR & WebQSP & ELQ            & ELQ  & 94.6 & 0.49 \\
CR & WebQSP & NSM            & ELQ  & 28.7 & 1.65 \\
CR & WebQSP & GraftNet       & GEM  & 58.2 & 36.55 \\
CR & WebQSP & ELQ+cosine     & ELQ  & 48.6 & 22.33 \\
\midrule
RS & CWQ    & ELQ            & ELQ  & 74.0 & 20.31 \\
RS & CWQ    & GraftNet-Orig  & GEM  & 68.6 & 11.24 \\
RS & CWQ    & NSM            & ELQ  & 43.2 & 8.33 \\
RS & CWQ    & GraftNet       & GEM  & 23.2 & 13.71 \\
\midrule
RS & WebQSP & ELQ            & ELQ  & 94.7 & 50.95 \\
RS & WebQSP & GraftNet-Orig  & GEM  & 89.3 & 41.67 \\
RS & WebQSP & NSM            & ELQ  & 28.2 & 15.07 \\
RS & WebQSP & GraftNet       & GEM  & 38.2 & 28.43 \\
\midrule
\multicolumn{6}{l}{\textit{Oracle-seeded reference (NSM-GEM); not included in the 24-cell Spearman correlation:}} \\
CR & CWQ    & NSM-GEM        & GEM  & 60.8  & 30.25 \\
CR & WebQSP & NSM-GEM        & GEM  & 36.24 & 25.56 \\
RS & CWQ    & NSM-GEM        & GEM  & 67.01 & 37.24 \\
RS & WebQSP & NSM-GEM        & GEM  & 35.63 & 25.08 \\

\bottomrule
\end{tabular}
\end{table*}

\section{Failure Analysis: Case Examples and PPR Visualisation}\label{app:failure_cases}

\subsection{Failure Case Examples}

The two cases below are drawn from RS (P7, relation synonym swap) evaluated
with GraftNet-Orig subgraphs, illustrating two distinct failure mechanisms.
RS failures are not monolithic: in Case~1 the gold answer is absent from
the retrieved subgraph; in Case~2 the answer is reachable but the GNN assigns
higher probability to an incorrect entity.

\paragraph{Case 1: Answer absent from the perturbed subgraph.}
The perturbed question retrieves a different subgraph whose entity set does
not include the gold answer MID.\@ Correct GNN reasoning is irrelevant because
the answer entity is not among the candidates.

\begin{table*}[ht]
\centering
\small
\caption{Case 1 examples: answer absent from perturbed subgraph.  The relation
  synonym shift changes the question embedding used to weight per-relation PPR
  walks, causing PPR to converge to a different subgraph that excludes the gold
  entity.  Hit@1\,=\,0 regardless of GNN reasoning quality.}\label{tab:failure_case1}
\setlength{\tabcolsep}{4pt}
\begin{tabular}{p{2.0cm} p{6.0cm} p{6.0cm}}
\toprule
 & \textbf{CWQ example} & \textbf{WebQSP example} \\
\midrule
Orig Q & \textit{What character was \textbf{played} by both Josh Pence and Armie Hammer?}
       & \textit{What kind of \textbf{money} do I bring to Mexico?} \\
Pert Q & \textit{What character was \textbf{portrayed} by both Josh Pence and Armie Hammer?}
       & \textit{What type of \textbf{currency} should I take to Mexico?} \\
Gold   & \texttt{m.09tb\_f3} (Tyler Winklevoss) & \texttt{m.012ts8} (Mexican peso) \\
Pred (orig) & \texttt{m.09tb\_f3} \cmark{} & \texttt{m.012ts8} \cmark{} \\
Pred (pert) & \texttt{m.0hyf5kr} \xmark{} & \texttt{m.04sqj} \xmark{} \\
Gold in pert SG & \textbf{No} & \textbf{No} \\
\bottomrule
\end{tabular}
\end{table*}

\paragraph{Case 2: Answer present but GNN reasoning fails.}
The gold answer entity is reachable in the perturbed subgraph, but GNN-RAG
assigns higher final probability to a different entity.  This case illustrates
that even when $\alpha_{\mathrm{pert}}$\,=\,1, the GNN can still fail if the
perturbed subgraph topology provides insufficient message-passing support for
the answer node.

\begin{table*}[ht]
\centering
\small
\caption{Case 2 examples: answer present in perturbed subgraph but GNN
  reasoning fails.  In the CWQ case, the perturbed subgraph has only 89 triples
  vs.\ 6,412 in the original; answer entities are present as isolated nodes with
  insufficient connectivity for GNN message-passing.  In the WebQSP case, voice
  shift alters the SBERT encoding, redirecting ReaRev's query-entity attention
  despite correct connectivity.}\label{tab:failure_case2}
\setlength{\tabcolsep}{4pt}
\begin{tabular}{p{2.0cm} p{6.0cm} p{6.0cm}}
\toprule
 & \textbf{CWQ example} & \textbf{WebQSP example} \\
\midrule
Orig Q & \textit{What countries \textbf{in} the Chamorro Time Zone are \textbf{in} Oceania?}
       & \textit{What district \textbf{does} Nancy Pelosi \textbf{represent}?} \\
Pert Q & \textit{What countries \textbf{within} the Chamorro Time Zone are \textbf{located in} Oceania?}
       & \textit{What district \textbf{is represented by} Nancy Pelosi?} \\
Gold   & \texttt{m.05cnr} (Guam), \texttt{m.034tl} (N.\ Mariana Is.) & \texttt{m.0b10j3} (CA districts) \\
Pred (orig) & \texttt{m.05cnr} \cmark{} & \texttt{m.0b10j3} \cmark{} \\
Pred (pert) & \texttt{m.0dff7s} \xmark{} & \texttt{m.09c7w0} (United States) \xmark{} \\
Gold in pert SG & \textbf{Yes (both)} & \textbf{Yes (all 3)} \\
Pert SG triples & 89 vs.\ 6{,}412 orig & 351 vs.\ 2{,}146 orig \\
\bottomrule
\end{tabular}
\end{table*}

\begin{table*}[ht]
\centering
\caption{Bonferroni 99.8\% CIs ($\alpha^{*} = 0.05/24 \approx 0.002$)
  for $\deltaAG$ EM, all perturbation types, both datasets.
  Attack codes: ES\,=\,entity alias; S1\,=\,rel.\ para.; S2\,=\,voice flip;
  CR\,=\,comp.\ restruct.; S3\,=\,ent.\ ins.; S4\,=\,mix.\ noise; RS\,=\,rel.\ syn.
  All five primary conclusions survive correction; effect sizes 10--52\,pp remain far from zero.
  Sub-1\,pp differences (e.g.\ GEM\,+\,Cosine vs.\ GEM\,+\,Flat) remain non-significant.}\label{tab:bonferroni}
\small
\setlength{\tabcolsep}{4pt}
\begin{tabular}{lrrrr}
\toprule
Attack & Dataset & Mean $\deltaAG$ & 95\% CI & 99.8\% CI \\
\midrule
ES & CWQ    & 0.526 & [0.509,\,0.543] & [0.501,\,0.553] \\
ES & WebQSP & 0.743 & [0.721,\,0.764] & [0.709,\,0.776] \\
\midrule
S1 & CWQ    & 0.527 & [0.511,\,0.543] & [0.501,\,0.552] \\
S1 & WebQSP & 0.739 & [0.718,\,0.761] & [0.707,\,0.773] \\
\midrule
S2 & CWQ    & 0.524 & [0.507,\,0.540] & [0.499,\,0.550] \\
S2 & WebQSP & 0.739 & [0.718,\,0.761] & [0.703,\,0.773] \\
\midrule
\textbf{CR} & CWQ    & \textbf{0.522} & \textbf{[0.506,\,0.539]} & \textbf{[0.496,\,0.547]} \\
\textbf{CR} & WebQSP & \textbf{0.738} & \textbf{[0.717,\,0.760]} & \textbf{[0.702,\,0.772]} \\
\midrule
S3 & CWQ    & 0.525 & [0.508,\,0.541] & [0.499,\,0.552] \\
S3 & WebQSP & 0.739 & [0.718,\,0.760] & [0.707,\,0.774] \\
\midrule
S4 & CWQ    & 0.523 & [0.506,\,0.540] & [0.497,\,0.549] \\
S4 & WebQSP & 0.738 & [0.716,\,0.759] & [0.705,\,0.772] \\
\midrule
\textbf{RS} & CWQ    & \textbf{0.326} & \textbf{[0.308,\,0.343]} & \textbf{[0.299,\,0.354]} \\
\textbf{RS} & WebQSP & \textbf{0.234} & \textbf{[0.210,\,0.257]} & \textbf{[0.197,\,0.272]} \\
\bottomrule
\end{tabular}
\end{table*}

\subsection{PPR Mass Topology Mismatch: Example Visualisation}\label{app:ppr_vis}

\begin{figure*}[t]
\centering
\includegraphics[width=\linewidth]{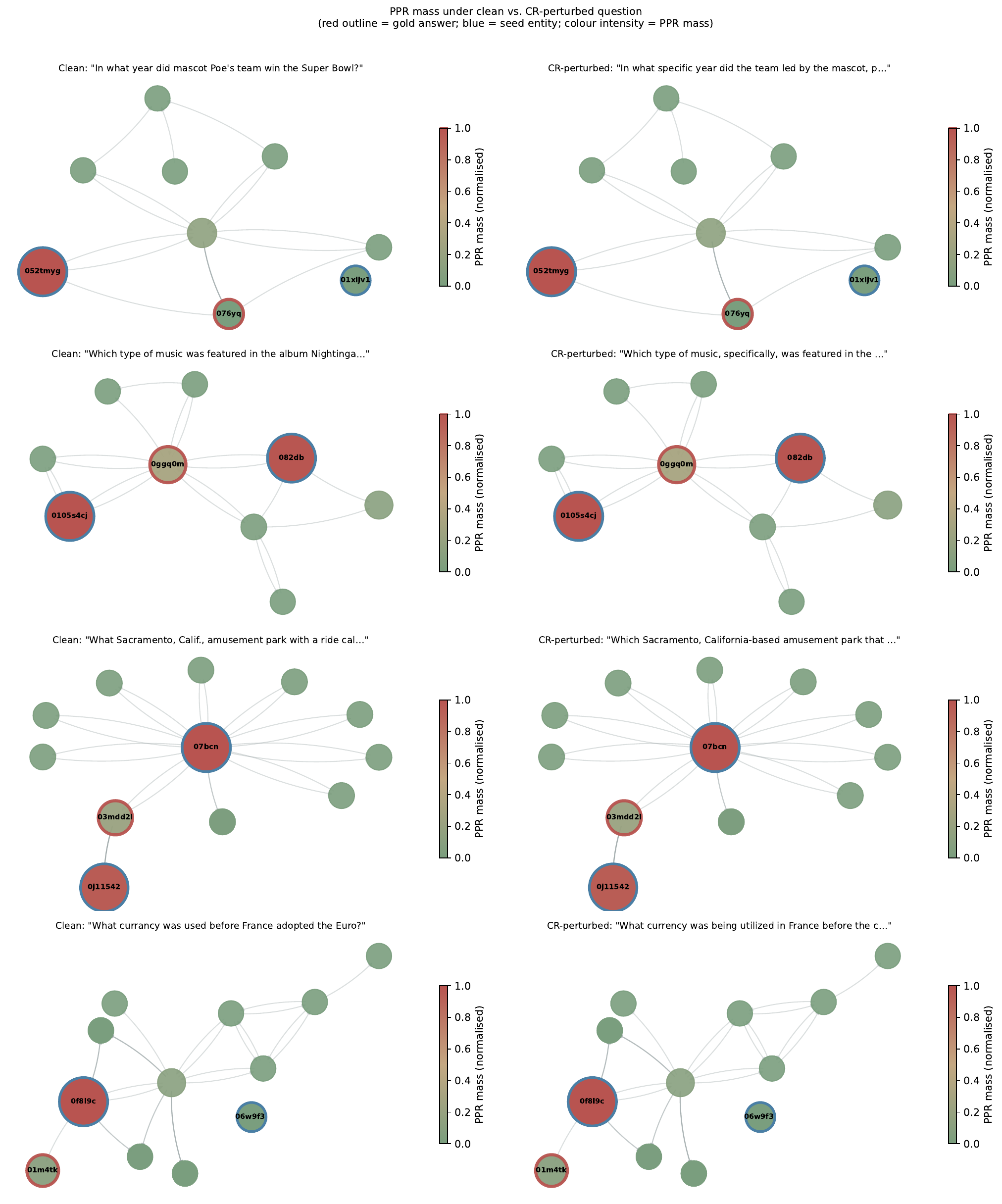}
\caption{\textbf{Answer present but unreachable under CR: four examples.}
  Each row shows a clean/perturbed question pair; node colour $\propto$ PPR mass
  (red\,=\,high, green\,=\,low); blue double-circle\,=\,ELQ seed; red double-circle\,=\,gold answer.
  The gold answer node is \emph{present} in both subgraphs with near-identical PPR mass,
  yet exact-match accuracy collapses from 1 (clean) to 0 (CR).
  ELQ reuses the same seed (Trip.\ Jac.\,=\,0.885 CWQ CR), but the PPR walk is anchored
  to the original hop order: paths required by the \emph{restructured} reasoning chain are
  not emphasised. The GNN decoder is not the culprit: on the clean subgraph with a CR
  question it achieves 52.76\% CWQ EM (Table~\ref{tab:rearev_hits}).
  This is the \emph{answer presence $\neq$ answer reachability} finding
  (74\% of CR failures have the gold answer in the subgraph; EM\,=\,0.68\%).}
\label{fig:ppr-mismatch}
\end{figure*}

Figure~\ref{fig:ppr-mismatch} shows four examples of how compositional
restructuring (CR) exposes the topology-anchoring failure in GNN-RAG.
Each row shows the same 2-hop entity neighbourhood under the clean
question (left) and the CR-perturbed question with hop order reversed (right).
Node colour encodes PPR mass (red = high mass, green = low mass);
blue double-circles are ELQ-linked seeds; the red double-bordered node is the gold answer.

ELQ's PPR walk is anchored to the original question's reasoning chain;
the hop-reversed path from the seed entity to the gold answer is not
emphasised in the perturbed subgraph, so the GNN cannot navigate to the
answer via the restructured traversal order.
The bottleneck is the subgraph topology, not the GNN decoder: when the decoder
receives the \emph{clean} subgraph with the perturbed question, it achieves
52.76\% CWQ EM (fast-mode; instruction cosine\,=\,0.915, Table~\ref{tab:rearev_hits}),
confirming that the decoder handles CR well when path topology is correct.

\section{Secondary Perturbation Types and Ablations}\label{app:ablation_full}
\subsection{ES Answerability Confound}\label{app:answerability}

45.6\% of CWQ ES questions are unanswerable by construction: the substituted
entity has no Freebase path to any valid answer for the question's relation
chain.  CR and RS are ${\geq}99.8\%$ answerable.
Table~\ref{tab:es_answerable_v1} (Appendix~\ref{app:es_diagnostic}) separates
the two effects: GraftNet-BFS recovers 14.1\% CWQ EM on answerable questions
(vs.\ 8.2\% overall), confirming that structural unanswerability suppresses the
overall ES number independently of EL failure. ELQ remains near zero regardless
of answerability because wrong seeds completely derail PPR.

\subsection{Subgraph Component Ablation: Seeds \texorpdfstring{\texttimes}{x} PPR} \label{app:sgablation}

Table~\ref{tab:sgablation} decouples seed quality (GEM oracle vs.\ ELQ
production) from PPR flavour (cosine question-embedding weighting vs.\ flat
uniform weighting).  The key finding is that \emph{seed quality dominates}: GEM
seeds account for ${\approx}15$\,pp of the CR improvement regardless of PPR
type.  The fully deployment-realistic configuration (ELQ seeds\,+\,cosine PPR)
reaches 14.98\% CWQ CR, which is a meaningful gain over the 0.68\% ELQ flat-PPR
baseline, but well below the GEM-seeded upper bound of 29.82\%.

\textbf{Significance of the GEM\,+\,Cosine vs.\ GEM\,+\,Flat difference.}
On CWQ CR, the 0.46\,pp gap (29.79\% vs.\ 30.25\%) is not statistically
significant: paired bootstrap test ($n_{\mathrm{boot}}{=}10{,}000$), $p = 0.478$.
This confirms that continuous cosine PPR weighting provides no benefit over
uniform flat PPR on CWQ once seed quality is fixed: the performance ceiling
is determined by seed quality, not PPR flavour.
On WebQSP CR, GEM\,+\,Cosine is 5.6\,pp higher (31.18\% vs.\ 25.56\%;
$p < 0.001$), indicating that cosine weighting does help on shorter
1--2 hop chains where embedding alignment is more decisive.
The deployment-realistic gap (ELQ\,+\,Cosine vs.\ ELQ\,+\,Flat on CWQ: 14.98\% vs.\
15.38\%) is similarly non-significant ($p = 0.58$), confirming that \emph{the bottleneck
for deployment is seed quality, not PPR weighting}.
The GraftNet-V2 binary-mask variant (Appendix~\ref{app:graftnetv2}) provides
complementary evidence.

\textbf{Reframing.}
The non-significant CWQ difference and the significant WebQSP difference together
support a refined conclusion: continuous cosine PPR weighting is not the driver
of GraftNet's CR improvement on CWQ (where the $p$-value confirms a tie), but
it contributes on WebQSP's simpler hop structure.
A deployment-realistic improvement to the CR bottleneck on CWQ therefore
requires better entity linking (better seeds), not a more sophisticated PPR variant.

\begin{table*}[ht]
\centering
\caption{Seeds\,$\times$\,PPR ablation: GNN-RAG EM (\%) for CR and RS.
  GEM\,=\,gold SPARQL MIDs (oracle); ELQ\,=\,production linker.
  Clean baselines: CWQ\,=\,52.9\%, WebQSP\,=\,74.3\%.
  The GEM/ELQ gap (${\approx}15$\,pp for CR) quantifies the seed-quality ceiling;
  cosine PPR adds marginal benefit on top.
  ``Main paper GraftNet (fixed)'' uses GloVe relation embeddings (per relation),
  whereas the controlled GEM\,+\,Cosine row uses sentence-level question embeddings;
  this granularity accounts for the RS gap (13.71\% vs.\ 9.86\%) while CR is
  unaffected (29.82\% vs.\ 29.79\%).}
\label{tab:sgablation}
\small
\setlength{\tabcolsep}{4pt}
\begin{tabular}{llrrrr}
\toprule
 & & \multicolumn{2}{c}{\textbf{CWQ}} & \multicolumn{2}{c}{\textbf{WebQSP}} \\
\cmidrule(lr){3-4}\cmidrule(lr){5-6}
Seeds & PPR & CR (\%) & RS (\%) & CR (\%) & RS (\%) \\
\midrule
GEM   & Cosine (question-emb.) & \textbf{29.79} & 9.86           & \textbf{31.18} & 28.31 \\
GEM   & Flat (uniform)        & 30.25          & \textbf{37.24} & 25.56          & \textbf{25.08} \\
ELQ   & Cosine (question-emb.) & 14.98          & 9.92           & 22.33          & 28.31 \\
ELQ   & Flat (uniform)        & 15.38          & 8.73           & 17.69          & 25.69 \\
\midrule
\multicolumn{2}{l}{\textit{Main paper GraftNet (fixed)}} & \textit{29.82} & \textit{13.71} & \textit{36.55} & \textit{28.43} \\
\multicolumn{2}{l}{\textit{ELQ (production baseline)}}  & \textit{0.68}  & \textit{20.31} & \textit{0.49}  & \textit{50.95} \\
\bottomrule
\end{tabular}
\end{table*}

\subsection{Secondary Perturbation Types}\label{app:ablation}

Table~\ref{tab:secondary} reports GNN-RAG EM under the four secondary
perturbation types (S1--S4).  All four collapse performance to ${\leq}0.65\%$
EM, comparable in magnitude to CR.  Critically, unlike CR and RS, the secondary
types simultaneously stress EL and downstream stages: S1 alters entity surface
forms (SeedHit drops to 44\% on CWQ, Table~\ref{tab:el}), making clean
stage-isolation impossible.  S2 (active\,$\leftrightarrow$\,passive) mostly
preserves entity mentions but voice changes are less semantically targeted.
S3 and S4 insert adversarial entities or apply mixed noise, both of which
partially perturb ELQ seeds.  Because these types conflate EL failure with
downstream failure, they are reported for completeness rather than as
mechanistically interpretable probes.  A consolidated view across all seven
perturbation types under ELQ subgraphs is in Section~\ref{app:all_pert}.

\begin{table*}[ht]
\centering
\caption{GNN-RAG EM under secondary perturbation types (ELQ seeds;
  bootstrap 95\% CI on $\deltaAG$, $n{=}1000$; per-comparison, uncorrected).
  All four types collapse EM to ${\leq}0.65\%$, comparable to CR (0.68\%),
  but unlike CR they co-stress EL and downstream stages simultaneously,
  preventing clean stage-isolation.}
\label{tab:secondary}
\small
\setlength{\tabcolsep}{4pt}
\begin{tabular}{lrrrr}
\toprule
 & \multicolumn{2}{c}{\textbf{CWQ}} & \multicolumn{2}{c}{\textbf{WebQSP}} \\
\cmidrule(lr){2-3}\cmidrule(lr){4-5}
Type & EM (\%) & $\deltaAG$ {(pp)} [95\% CI] & EM (\%) & $\deltaAG$ {(pp)} [95\% CI] \\
\midrule
S1 (rel.\ paraphrase)    & 0.20 & 52.7 [50.9, 54.3] & 0.43 & 73.9 [71.8, 76.1] \\
S2 (voice flip)          & 0.51 & 52.4 [50.8, 54.0] & 0.37 & 73.9 [72.0, 76.1] \\
S3 (entity insertion)    & 0.40 & 52.5 [51.0, 54.3] & 0.43 & 73.9 [71.9, 76.0] \\
S4 (mixed lexical noise) & 0.57 & 52.3 [50.7, 53.9] & 0.49 & 73.8 [71.8, 76.0] \\
\bottomrule
\end{tabular}
\end{table*}

\paragraph{S1--S4 with GEM oracle subgraphs.}
Table~\ref{tab:secondary_gem} repeats the S1--S4 evaluation with GEM oracle seeds
(NSM-GEM and GraftNet-GEM variants), isolating the effect of subgraph quality from
EL failure.  Under oracle subgraphs, all four secondary types recover to
${\approx}30$\,\% CWQ EM, matching CR with the same oracle seeds (NSM-GEM: 30.25\%,
GraftNet-GEM: 29.79\%).
A caveat applies to S1: because S1 alters entity surface forms (SeedHit drops to
44\% on CWQ, Table~\ref{tab:el}), GEM oracle seeds simultaneously fix both the EL
failure and subgraph quality, so S1's recovery reflects a combined EL+subgraph fix
rather than a pure subgraph isolation.  For S2--S4, entity mentions are better
preserved, so oracle seeds function closer to a genuine subgraph-only fix.
Excluding S1, the uniform recovery for S2--S4 provides cleaner support that
\emph{subgraph quality is the shared bottleneck}.

\begin{table*}[ht]
\centering
\caption{GNN-RAG EM under secondary perturbation types with GEM oracle seeds
  (bootstrap 95\% CI on $\deltaAG$, $n{=}1000$, uncorrected).
  All four types recover to ${\approx}30\%$ CWQ / ${\approx}26$--32\% WebQSP EM,
  matching the CR-with-oracle baseline and confirming subgraph quality as the
  shared bottleneck. Clean EM: CWQ\,=\,52.9\%, WebQSP\,=\,74.3\%.}
\label{tab:secondary_gem}
\small
\setlength{\tabcolsep}{4pt}
\begin{tabular}{llrrrr}
\toprule
 & & \multicolumn{2}{c}{\textbf{CWQ}} & \multicolumn{2}{c}{\textbf{WebQSP}} \\
\cmidrule(lr){3-4}\cmidrule(lr){5-6}
Subgraph & Type & EM (\%) & $\deltaAG$ {(pp)} [95\% CI] & EM (\%) & $\deltaAG$ {(pp)} [95\% CI] \\
\midrule
\multicolumn{6}{l}{\textit{NSM-GEM (oracle seeds, uniform PPR)}} \\
 & S1 (rel.\ paraphrase)    & 30.93 & 22.0 [20.2,\,23.8] & 26.11 & 48.1 [45.6,\,50.8] \\
 & S2 (voice flip)          & 30.42 & 22.5 [20.8,\,24.2] & 26.54 & 47.8 [45.1,\,50.4] \\
 & S3 (entity insertion)    & 30.56 & 22.3 [20.6,\,24.1] & 25.56 & 48.7 [46.0,\,51.4] \\
 & S4 (mixed lexical noise) & 30.59 & 22.3 [20.6,\,24.0] & 25.99 & 48.3 [45.6,\,50.9] \\
 & \textit{CR (reference)}  & \textit{30.25} & \textit{0.227} & \textit{25.56} & \textit{0.487} \\
\midrule
\multicolumn{6}{l}{\textit{GraftNet-GEM (oracle seeds, question-embedding PPR)}} \\
 & S1 (rel.\ paraphrase)    & 30.30 & 22.6 [20.8,\,24.4] & 31.54 & 42.7 [40.0,\,45.4] \\
 & S2 (voice flip)          & 29.40 & 23.5 [21.6,\,25.5] & 31.60 & 42.7 [40.0,\,45.5] \\
 & S3 (entity insertion)    & 29.91 & 23.0 [21.1,\,24.8] & 31.18 & 43.1 [40.3,\,46.0] \\
 & S4 (mixed lexical noise) & 30.08 & 22.8 [21.0,\,24.6] & 31.24 & 43.1 [40.3,\,45.9] \\
 & \textit{CR (reference)}  & \textit{29.79} & \textit{0.231} & \textit{31.18} & \textit{0.431} \\
\bottomrule
\end{tabular}
\end{table*}

\subsection{All Perturbation Types: ELQ Baseline}\label{app:all_pert}

Table~\ref{tab:all_pert} provides a consolidated view across all seven
perturbation types evaluated with ELQ subgraphs.  The table reveals a striking
asymmetry: RS (P7) is uniquely robust (20.31\% CWQ, 50.95\% WebQSP) while all
other types collapse to ${\leq}0.7\%$. The explanation is structural: for RS,
ELQ's entity seeds are identical to the clean question's, so the PPR walk
effectively replays the original subgraph, preserving $\alpha_{\mathrm{pert}}$
at 74.0\%.  For every other type, either entity seeds change (ES, S1, S3, S4)
or the subgraph topology diverges from the restructured question (CR), where both
paths lead to near-zero EM.  This consolidation view motivates why RS and CR
were selected as the two primary attacks: RS isolates the GNN decoder while CR
isolates the subgraph construction bottleneck.

\begin{table*}[ht]
\centering
\caption{GNN-RAG EM (\%) with ELQ subgraph, all perturbation types.
  Clean: CWQ\,=\,52.9\%, WebQSP\,=\,74.3\%.
  RS (P7) is uniquely robust because ELQ reuses the original subgraph
  unchanged; all other types cause near-total collapse.
  Bootstrap 95\% CIs are per-comparison and uncorrected for multiple
  comparisons; Bonferroni 99.8\% CIs in
  Appendix~\ref{app:bonferroni}.}\label{tab:all_pert}
\small
\setlength{\tabcolsep}{4pt}
\begin{tabular}{llrrrr}
\toprule
Code & Type & CWQ EM & CWQ $\deltaAG$ {(pp)} [95\% CI] & WSP EM & WSP $\deltaAG$ {(pp)} [95\% CI] \\
\midrule
ES (P1)       & Entity alias      & 0.31  & 52.6 [51.0,\,54.3] & 0.06  & 74.3 [71.9,\,76.5] \\
S1 (P2)       & Rel.\ paraphrase  & 0.20  & 52.7 [50.9,\,54.3] & 0.43  & 73.9 [71.8,\,76.1] \\
S2 (P3)       & Voice flip        & 0.51  & 52.4 [50.8,\,54.0] & 0.37  & 73.9 [72.0,\,76.1] \\
\textbf{CR (P4)} & \textbf{Comp.\ restruct.} & \textbf{0.68} & \textbf{52.2 [50.5,\,53.8]} & \textbf{0.49} & \textbf{73.8 [71.6,\,75.8]} \\
S3 (P5)       & Entity insertion  & 0.40  & 52.5 [51.0,\,54.3] & 0.43  & 73.9 [71.9,\,76.0] \\
S4 (P6)       & Mixed noise       & 0.57  & 52.3 [50.7,\,53.9] & 0.49  & 73.8 [71.8,\,76.0] \\
\textbf{RS (P7)} & \textbf{Rel.\ synonym}   & \textbf{20.31} & \textbf{32.6 [30.7,\,34.4]} & \textbf{50.95} & \textbf{23.4 [21.0,\,25.7]} \\
\bottomrule
\end{tabular}
\end{table*}

\subsection{GMT-KBQA and EPR-KGQA as Architectural Controls}\label{app:gmt}

These two systems serve as architectural controls that help isolate whether the
observed vulnerability is specific to PPR-based subgraph retrieval or a general
property of GNN-KGQA pipelines.

\paragraph{GMT-KBQA.}
GMT-KBQA~\cite{das2021case} generates S-expressions directly from the question,
bypassing the EL-then-subgraph-retrieval pipeline.  Table~\ref{tab:gmt} shows
that structural and paraphrase perturbations (CR-type, RS-type, S2-type) cause
only ${\leq}5$\,pp degradation, compared to GNN-RAG's 52\,pp CR collapse.
Entity-related perturbations (ES-type, CR-type with entity insertion) cause
18--26\,pp degradation because S-expression generation is sensitive to entity
surface forms.  The contrast confirms that the GNN-RAG vulnerability to CR is
specific to its fixed-subgraph retrieval stage: once that stage is bypassed,
compositional restructuring becomes near-harmless.

\begin{table*}[ht]
\centering
\caption{GMT-KBQA S-expression EM under perturbation.
  Clean: CWQ\,=\,36.9\%, WebQSP\,=\,35.5\%.
  Structural perturbations (CR-type, RS-type, S2-type) cause ${\leq}5$\,pp drop,
  confirming S-expression generation is robust to compositional restructuring.
  Entity-type perturbations cause 18--26\,pp degradation,
  matching GNN-RAG's EL vulnerability profile.}\label{tab:gmt}
\small
\setlength{\tabcolsep}{5pt}
\begin{tabular}{llrrrr}
\toprule
Perturbation & Type & \multicolumn{2}{c}{CWQ} & \multicolumn{2}{c}{WebQSP} \\
\cmidrule(lr){3-4}\cmidrule(lr){5-6}
 & & EM (\%) & Drop & EM (\%) & Drop \\
\midrule
Entity alias subst.  & ES-type  & 18.3 & $-$18.6 & 10.8 & $-$24.7 \\
Adversarial entity   & CR-type  & 18.5 & $-$18.4 & 9.4  & $-$26.1 \\
Question reordering  & CR-type  & 31.9 & $-$5.0  & 35.9 & $+$0.4  \\
Relation synonym     & RS-type  & 32.3 & $-$4.6  & 32.0 & $-$3.5  \\
Active-passive flip  & S2-type  & 34.9 & $-$2.0  & 35.7 & $+$0.2  \\
\bottomrule
\end{tabular}
\end{table*}

\paragraph{EPR-KGQA.}
EPR-KGQA~\cite{EPR-KGQA} uses atomic adjacency patterns (entity-relation-entity
tuples) indexed offline and selected at query time, replacing PPR-based retrieval
entirely.  Under CR, EPR-KGQA retains 59.22\% CWQ and 64.25\% WebQSP Hit@1
(vs.\ 0.68\% and 0.49\% for GNN-RAG ELQ), a near-baseline result despite the
compositional restructuring.  Under RS, EPR-KGQA retains 59.76\% CWQ and 63.09\%
WebQSP.  The near-zero degradation under both attacks shows that \emph{single-shot}
pattern-matching retrieval is robust to both attack types.
Contrast this with ExplaiGNN below: EPR-KGQA's robustness is not attributable
solely to PPR avoidance, but to its single-shot retrieval that does not propagate
errors across turns.

\textbf{Metric note.}
EPR-KGQA and GNN-RAG use the same MID-indexed Hit@1.
EPR-KGQA's NSM backbone indexes answers by Freebase MID (not entity name):
\texttt{basic\_dataset.py} selects \texttt{kb\_id} when \texttt{answer[`kb\_id']}
is a string (which it is for CWQ/WebQSP), and GNN-RAG's evaluator operates
on integer indices into the same MID-keyed entity vocabulary
(\texttt{entity2name=None} for non-\texttt{sr-} datasets).
Both systems therefore report top-1 accuracy over the same gold-MID entity set.
The ${\sim}58$\,pp gap under CR is a direct architectural comparison with
no metric conversion needed; it is not an artefact of metric leniency.

\paragraph{ExplaiGNN.}
ExplaiGNN~\cite{explaignn2022} is a conversational KGQA system (Wikidata-based)
that processes multi-turn question sequences using a GNN over iteratively refined
subgraphs.  Unlike GNN-RAG's single-question PPR retrieval, ExplaiGNN retrieves
subgraphs via entity-relation pattern matching similar to EPR-KGQA, chaining
evidence across turns.  We evaluate the published checkpoint on the ConvMix test
set (4,800 turns) under CR and RS perturbations applied to each conversation turn.

Table~\ref{tab:explaignn} shows results.
Clean P@1 is 33.9\%, consistent with the originally reported ConvMix result.
Under CR, P@1 drops to 9.8\% ($-$24.1\,pp; $-$71\% relative); under RS, P@1
drops to 10.2\% ($-$23.7\,pp; $-$70\% relative).  Both attacks cause a severe
collapse comparable in magnitude to GNN-RAG ELQ under CR (0.68\% CWQ,
$-$52\,pp).  The ExplaiGNN result shows that iterative subgraph reasoning systems
are \emph{not} immune: when the conversational context is compositionally
restructured or relation synonyms are swapped, the pattern-matching retrieval
fails to chain the correct entities across turns, and performance collapses.
The contrast with EPR-KGQA's near-baseline retention (59.2\% P@1 under CR)
suggests the vulnerability in ExplaiGNN arises from the iterative,
turn-dependent retrieval: each turn conditions on the previous turn's subgraph,
so a single perturbed turn corrupts all downstream reasoning.

\textbf{Scope note.}
This comparison is qualitative only.
ExplaiGNN uses a different KG (Wikidata vs.\ Freebase), benchmark (ConvMix
vs.\ CWQ/WebQSP), task type (multi-turn conversational vs.\ single-question),
and metric (P@1 vs.\ EM); the collapse magnitudes ($>$70\% relative) are
directionally consistent with the GNN-RAG finding.

\begin{table}[ht]
\centering
\caption{ExplaiGNN P@1, MRR {(Mean Reciprocal Rank)}, and {hits@5 (H@5)} on ConvMix test set (4,800 turns) under
  CR and RS perturbations.  Clean\,=\,published checkpoint, unperturbed.
  Both attacks cause severe collapse ($>$70\% relative P@1 drop), showing that
  iterative subgraph-chaining systems are vulnerable to the same perturbation
  types as single-question GNN-RAG, despite using pattern-matching retrieval.}
\label{tab:explaignn}
\small
\setlength{\tabcolsep}{5pt}
\begin{tabular}{lrrr}
\toprule
Condition & P@1 (\%) & MRR & H@5 (\%) \\
\midrule
Clean              & 33.9 & 0.398 & 47.7 \\
CR (perturbed)     &  9.8 & 0.116 & 14.0 \\
RS (perturbed)     & 10.2 & 0.123 & 14.9 \\
\midrule
\textit{Gold-answer upper bound} & \textit{40.6} & \textit{0.471} & \textit{56.1} \\
\bottomrule
\end{tabular}
\end{table}

\subsection{GraftNet-V2: Binary Relation Mask (Diagnostic)}\label{app:graftnetv2}

GraftNet-V2 replaces continuous cosine weighting with a hard binary mask
(gold relations\,=\,1, others\,=\,0), eliminating embedding computation at
pruning time.  The three GraftNet variants share question-aware PPR but differ
in how relation scores are assigned: GraftNet-Orig (GloVe continuous), GraftNet
(mean-pooled word embeddings, continuous), GraftNet-V2 (binary).
From the Seeds\,$\times$\,PPR ablation (Table~\ref{tab:sgablation}), the
GEM\,+\,Flat configuration (equivalent to GraftNet-V2's binary masking with
correct seeds) achieves 30.25\% CWQ CR, statistically indistinguishable from
GEM\,+\,Cosine (29.82\%; paired bootstrap $p = 0.478$).
On CWQ, the performance ceiling is determined by seed quality, not PPR flavour.
On WebQSP, cosine weighting yields a significant 5.6\,pp gain over flat PPR
($p < 0.001$), suggesting the binary mask is a useful simplification only for
the longer-hop CWQ setting.

\subsection{RoG+PathOnly: Path Injection Methodology}\label{app:rog_methodology}

Predicted relation paths come from RoG's relation-path predictor, trained on
clean CWQ and WebQSP training data.  At inference, the predicted path (a
sequence of Freebase relation types) is prepended to the RoG prompt as a
structured context field before the question text.  RoG+PathOnly uses the
base GCR-Llama-2-7b checkpoint fine-tuned on clean question-answer pairs;
it differs from RoG+RA only in training data (RoG+RA also sees
relation-path-augmented examples during fine-tuning), not in inference
architecture or path injection format.  The near-identical CWQ results
(RoG+PathOnly 51.43\% vs.\ RoG+RA 50.47\% CR; 51.29\% vs.\ 51.91\% RS)
confirm that path injection at inference, not the augmented training signal,
is the dominant contributor to the accuracy gain.



\section{MetaQA: Stage-Isolation Results}\label{app:metaqa}

\paragraph{Setup.}
{To test whether the failure modes identified on Freebase are
schema-specific, we replicate the full protocol on
MetaQA~\cite{zhang2018variational}, a multi-hop KGQA benchmark over the
WikiMovies knowledge base (43{,}234 entities, 9 relations,
${\approx}134$K triples).  MetaQA is not derived from Freebase, and its
retrieval stage expands a seed neighbourhood rather than constructing a
PPR-weighted subgraph, making it a direct test of whether the
topology-anchoring failure is a property of the retrieval algorithm or of the
underlying KG.\@  We train the ReaRev backbone from scratch on the 2-hop and
3-hop splits, as no public MetaQA checkpoint accompanies GNN-RAG, reaching
clean Hit@1 of 99.74\% and 95.88\% respectively; the GNN-RAG authors report
98.6\% on MetaQA-3 for reference.  All seven perturbation types are
regenerated with the same Llama-3.3-70B-Instruct prompt templates
(Figures~\ref{fig:all_prompts} and~\ref{fig:secondary_prompts}) and the same
validity gates (Appendix~\ref{app:perturbation_rules}); only the knowledge
base changes.  Answer preservation is verified directly against the MetaQA
triple store, which replaces the SPARQL denotation check used for Freebase.
Evaluation covers the complete test sets (14{,}872 questions on 2-hop;
14{,}274 on 3-hop).}

\paragraph{No collapse under any perturbation type.}
{Table~\ref{tab:metaqa} reports Hit@1 for all seven types.  The most
damaging attack is S1 (relation paraphrase), costing 13.09\,pp on 3-hop and
12.97\,pp on 2-hop; every other type stays within 7.2\,pp of the clean
baseline.  The contrast with Freebase is stark: the CR attack that drives
GNN-RAG from 52.9\% to 0.68\% EM on CWQ, a 52.22\,pp collapse
(Table~\ref{tab:main}), costs only 6.91\,pp on MetaQA-3 and 0.11\,pp on
MetaQA-2.  Because the perturbation generator, the validity filters, and the
GNN architecture are held fixed across the two settings, the difference is
attributable to the retrieval stage: MetaQA subgraphs are built by
neighbourhood expansion from the topic entity, so a restructured question
cannot misdirect a PPR walk that is never computed.  This is precisely the
prediction the stage-isolation analysis makes (Section~\ref{sec:analysis}),
tested on an independent knowledge base.  The ordering across types is also
consistent with the main results: the relation-side attacks (S1, RS) are the
most damaging, matching the finding that the instruction decoder is the
sensitive component once the subgraph is intact, whereas S3 (adversarial
entity insertion) is nearly harmless (${\leq}0.85$\,pp) when seeds are
supplied.}

\begin{table}[t]
\centering
\caption{{GNN-RAG Hit@1 (\%) on MetaQA under all seven perturbation types,
  full test sets (2-hop $n{=}14{,}872$; 3-hop $n{=}14{,}274$).
  Clean baselines: 3-hop\,=\,95.88, 2-hop\,=\,99.74.
  $\Delta$ is the drop from the corresponding clean baseline in percentage
  points.  Attack codes follow Table~\ref{tab:all_pert}: ES\,=\,entity alias,
  S1\,=\,relation paraphrase, S2\,=\,voice flip, CR\,=\,compositional
  restructuring, S3\,=\,entity insertion, S4\,=\,mixed noise,
  RS\,=\,relation synonym.
  No type produces the near-total collapse observed on Freebase
  under CR (Table~\ref{tab:main}).}}
\label{tab:metaqa}
\small
\setlength{\tabcolsep}{4pt}
\begin{tabular}{lrrrr}
\toprule
 & \multicolumn{2}{c}{\textbf{MetaQA-3}} & \multicolumn{2}{c}{\textbf{MetaQA-2}} \\
\cmidrule(lr){2-3}\cmidrule(lr){4-5}
Attack & Hit@1 & $\Delta$ (pp) & Hit@1 & $\Delta$ (pp) \\
\midrule
ES              & 88.97 & 6.91  & 96.61 & 3.13 \\
\textbf{S1}     & \textbf{82.79} & \textbf{13.09} & \textbf{86.77} & \textbf{12.97} \\
S2              & 92.05 & 3.83  & 98.13 & 1.61 \\
CR              & 88.97 & 6.91  & 99.63 & 0.11 \\
S3              & 95.03 & 0.85  & 99.35 & 0.39 \\
S4              & 88.76 & 7.12  & 95.81 & 3.93 \\
RS              & 90.94 & 4.94  & 97.40 & 2.34 \\
\bottomrule
\end{tabular}
\end{table}

\paragraph{End-to-end re-execution confirms retrieval invariance.}
{The results above hold the retrieved subgraph fixed and replace only the
question text, isolating the decoder.  To verify that retrieval is genuinely
invariant rather than merely held constant by construction, we additionally
re-run the \emph{entire} pipeline on the perturbed questions: entity linking,
subgraph construction, and GNN inference are all recomputed from the perturbed
text, under two seeding conditions (an automatic longest-match linker, and the
gold topic-entity annotations that MetaQA ships and that published systems in
this line consume).  Under CR, end-to-end Hit@1 changes by at most 0.6\,pp on
either hop relative to the same pipeline run on clean text.  Retrieval on
MetaQA is therefore not merely assumed stable but measured stable, which is
exactly the property that fails on Freebase, where CR leaves entity seeds
intact (97.5\% same-seed) yet redirects the PPR walk enough to erase
52.08\,pp of end-to-end accuracy.  The same experiment reproduces the
EL-conflation caveat of Appendix~\ref{app:ablation_full} on a second KB:
S3 degrades the automatic linker by 22.7\,pp (3-hop) and 18.6\,pp (2-hop)
because the injected distractor entity becomes the longest surface match, and
this propagates to an 8.3\,pp end-to-end drop on 2-hop; with the standard gold
seeds the same attack costs 0.4\,pp.  The apparent harmlessness of S3 in
Table~\ref{tab:metaqa} is thus a consequence of the seeding convention, not
evidence that distractor injection is benign for a deployed linker, exactly as
observed for S1--S4 on Freebase.}

{Table~\ref{tab:metaqa_e2e} reports the full end-to-end deltas for both
seeding conditions.  Two patterns are of note.  First, re-execution is
uniformly less damaging than the fixed-subgraph condition of
Table~\ref{tab:metaqa}.  The isolation condition confronts the decoder with a
subgraph built for the clean question, which is an adversarially hard mismatch,
whereas re-execution allows retrieval to rebuild a subgraph consistent with the
perturbed question, so that on a non-PPR retriever the mismatch largely
disappears.  The single exception is S1 on 2-hop, at $-12.1$\,pp under the
automatic linker and $-10.8$\,pp under gold seeds, where the relation
paraphrase degrades the instruction decoder even when the subgraph is
consistent.  This confirms that the attack is decoder-side rather than a
retrieval or linking artifact.  On 3-hop the same attack changes end-to-end
Hit@1 by only $-1.7$\,pp under the automatic linker and $-2.1$\,pp under gold
seeds.  Second, ES and CR share identical
fixed-subgraph deltas on 3-hop ($-6.91$\,pp in Table~\ref{tab:metaqa}) yet
diverge sharply under re-execution, ES improving by $11.3$\,pp because the
swapped entity retrieves an easier neighbourhood while CR changes by only
$-0.5$\,pp.  This divergence localises ES to the linking stage and CR to
neither retrieval nor decoding on this knowledge base.}

\begin{table}[t]
\centering
\caption{{GNN-RAG Hit@1 (\%) on MetaQA under full end-to-end re-execution,
  in which entity linking, retrieval, and GNN inference are all recomputed from
  the perturbed text, for the automatic longest-match linker (\textbf{auto})
  and the gold topic-entity seeds (\textbf{gold}).  $\Delta$ is the drop
  relative to the same-condition clean baseline.  Clean Hit@1: auto
  27.66/79.29, gold 30.69/80.45 (3-hop/2-hop).  Only within-condition deltas
  are comparable.  Compare to the fixed-subgraph condition of
  Table~\ref{tab:metaqa}.  Attack codes as in Table~\ref{tab:metaqa}.}}
\label{tab:metaqa_e2e}
\small
\setlength{\tabcolsep}{4pt}
\begin{tabular}{lrrrr}
\toprule
 & \multicolumn{2}{c}{\textbf{MetaQA-3} $\Delta$ (pp)} & \multicolumn{2}{c}{\textbf{MetaQA-2} $\Delta$ (pp)} \\
\cmidrule(lr){2-3}\cmidrule(lr){4-5}
Attack & auto & gold & auto & gold \\
\midrule
ES          & $+11.33$ & $+12.77$ & $-2.35$ & $-2.27$ \\
\textbf{S1} & $-1.66$ & $-2.06$ & $\mathbf{-12.14}$ & $\mathbf{-10.76}$ \\
S2          & $-1.21$ & $-0.90$ & $-4.02$ & $-1.70$ \\
CR          & $-0.53$ & $-0.46$ & $-0.28$ & $-0.02$ \\
S3          & $+0.28$ & $-0.05$ & $-8.34$ & $-0.32$ \\
S4          & $-1.12$ & $-1.17$ & $-6.30$ & $-3.78$ \\
RS          & $-0.57$ & $-0.70$ & $-2.07$ & $-1.98$ \\
\bottomrule
\end{tabular}
\end{table}

\paragraph{Scope.}
{Absolute accuracies in the end-to-end condition are bound to our reference
linker and retriever: clean gold-seeded Hit@1 is 30.69\% on 3-hop, well below
the 95.88\% obtained with the subgraphs shipped with the dataset.  Only
within-condition deltas should therefore be compared across rows, and we
report the clean-versus-perturbed difference under an identical operator
throughout.  ES is excluded from the end-to-end comparison because it changes
the gold answer by design (Section~\ref{sec:taxonomy}), so clean and perturbed
runs do not measure the same question.}

\section{Bonferroni Confidence Intervals}\label{app:bonferroni}

Table~\ref{tab:bonferroni} repeats the 14 $\deltaAG$ cells from
Table~\ref{tab:all_pert} (7 perturbation types $\times$ 2 datasets) with
Bonferroni 99.8\% intervals ($\alpha^{*} = 0.05/24 = 0.002$, dividing by
the 24 cells of the component ablation (Appendix~\ref{app:alpha_table});
$n{=}10{,}000$ bootstrap draws).

\noindent\textbf{Primary conclusions after Bonferroni adjustment.}
All five primary conclusions stated in Section~\ref{sec:conclusion} survive:
(a)~ELQ CR collapse ($\deltaAG \approx 0.522$, corrected CI lower bound 0.496);
(b)~GraftNet CR recovery (29.82\% vs.\ 0.68\%, far outside any CI width);
(c)~RS ELQ robustness ($\deltaAG \approx 0.326$ CWQ; corrected CI [0.299, 0.354]);
(d)~GNN decoder robustness (52.76\% $\mathrm{EM}_{\mathrm{fast}}$ with a clean subgraph);
(e)~EPR-KGQA near-baseline retention (effect sizes above 10\,pp, Table~\ref{tab:gmt}).
Corrected CI half-widths are approximately 2\,pp to 4\,pp, negligible relative to these effects.
The 0.46\,pp GEM\,+\,Cosine vs.\ GEM\,+\,Flat difference on CWQ CR does not survive
correction ($p = 0.478$), confirming seed quality, not PPR weighting, is the dominant
factor for CWQ; WebQSP shows a significant cosine benefit ($+5.6$\,pp, $p < 0.001$)
on shorter hop chains (Appendix~\ref{app:sgablation}).

\section{Licenses}\label{app:licenses}

All experimental assets are cited at first use and used consistently with
their intended research purposes.
\textit{Datasets:}
CWQ~\citep{talmor2018web} is available under Apache-2.0;
WebQSP~\citep{yih2016value} under the Microsoft Research License
(\url{https://www.microsoft.com/en-us/download/details.aspx?id=52763});
the Freebase 2015 RDF dump under CC-BY~2.5
(\url{https://developers.google.com/freebase}).
\textit{Evaluated systems:}
ELQ via BLINK~\citep{li2020efficient} is MIT-licensed;
EPR-KGQA~\citep{EPR-KGQA} Apache-2.0;
GMT-KBQA~\citep{das2021case} BSD-3-Clause;
GNN-RAG~\citep{gnnrag} carries no explicit license file at the time of
writing and is treated as research-use only.
\textit{Generation model:}
Llama-3.3-70B-Instruct~\citep{dubey2024llama} is released under the
Llama~3 Community License
(\url{https://huggingface.co/meta-llama/Llama-3.3-70B-Instruct}).
\textit{Released artefacts:}
Our perturbed datasets and evaluation code
(\url{https://anonymous.4open.science/r/atkgrag-E85C})
are released under CC~BY~4.0; downstream use must remain consistent
with the upstream research-only scope of CWQ, WebQSP, and Freebase.

\end{document}